\pdfoutput=1
\documentclass[11pt]{article}

\usepackage[preprint]{acl}
\usepackage{times}
\usepackage{latexsym}

\usepackage[T1]{fontenc}
\usepackage[utf8]{inputenc}

\usepackage{microtype}

\usepackage{inconsolata}

\usepackage{graphicx}
\usepackage{multicol}
\usepackage{tikz}
\usepackage{color, colortbl}
\usepackage{enumitem} % Add this to your preamble
\usepackage{amssymb}
\usepackage{makecell}
\usepackage{booktabs}
\usepackage{xcolor}
\usepackage{subcaption}
\usepackage{tikz}
\usepackage{pgffor}
\usepackage{tcolorbox}
\usepackage{multirow}
\usepackage[normalem]{ulem}
\usepackage{amsmath}
\usepackage{xparse}
\usepackage[linesnumbered,ruled,vlined]{algorithm2e}

\NewDocumentCommand\emojione{}{\raisebox{-0.75em}{\includegraphics[height=2.0em]{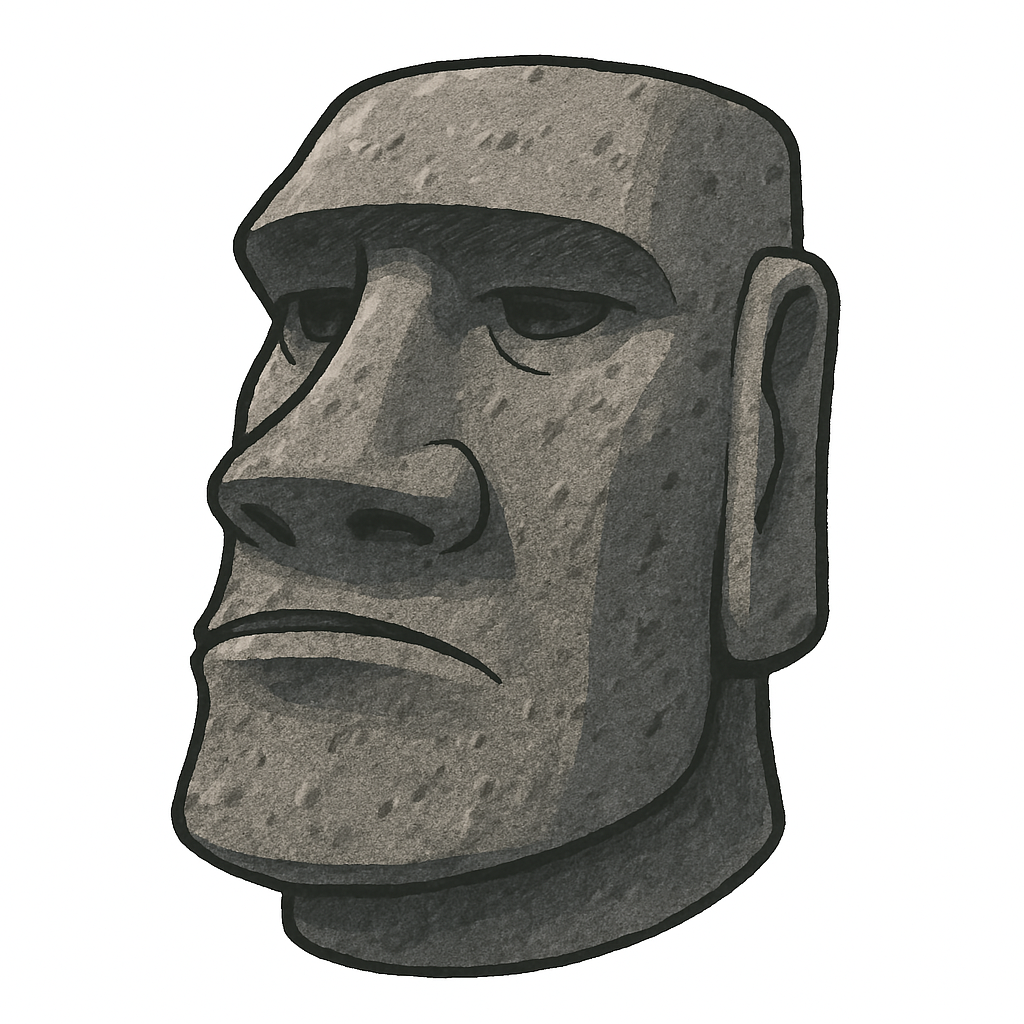}}}

\definecolor{lightgray}{gray}{0.9}
\usetikzlibrary{arrows.meta, positioning, shapes.geometric, shadows, backgrounds, fit, calc}

\title{\emojione\hspace{0.15em}\textsc{ConceptCarve}: Dynamic Realization of Evidence}

\author{Eylon Caplan \and Dan Goldwasser\\
  Purdue University, West Lafayette, IN, USA\\
  \texttt{\{ecaplan,dgoldwas\}@purdue.edu}\\}

\begin{document}
\maketitle
\begin{abstract} 
Finding evidence for human opinion and behavior at scale is a challenging task, often requiring an understanding of sophisticated thought patterns among vast online communities found on social media. For example, studying how \textit{gun ownership} is related to the perception of \textit{Freedom}, requires a retrieval system that can operate at scale over social media posts, while dealing with two key challenges: (1) identifying abstract concept instances, (2) which can be instantiated differently across different communities. To address these, we introduce \textsc{ConceptCarve}, an evidence retrieval framework that utilizes traditional retrievers and LLMs to dynamically characterize the search space during retrieval. Our experiments show that \textsc{ConceptCarve} surpasses traditional retrieval systems in finding evidence within a social media community. It also produces an interpretable representation of the evidence for that community, which we use to qualitatively analyze complex thought patterns that manifest differently across the communities.

\end{abstract}

\section{Introduction}
Human behavior and opinion are notoriously complex, particularly when mining textual resources in order to understand them \citep{kang-etal-2023-values, paulissen-wendt-2023-lauri, He_Wang_Zhao_Yang_2024}. If done well, a system that accurately quantifies human opinion at scale could supplement expensive polls and potentially reduce reliance on them. Likewise, it could demonstrate patterns in preferences like dietary habits~\cite{pilavr2021healthy,hashimoto2024meat} or vaccine hesitancy~\cite{zhang2023moral,qorib2023covid}. These patterns, which we refer to as \emph{trends}, capture the collective tendencies, attitudes, beliefs, or behaviors within a specific community or context. However, due to the complexity of human thought and expression, finding and quantifying supporting textual evidence of these trends is difficult, as it requires an understanding of what such evidence would look like in text (i.e. the \emph{realization} of the evidence).
\begin{figure}[!t]
    \centering
    \resizebox{\columnwidth}{!}{
    \def\vdist{2.7} % Adjust this value to change the spacing between boxes
    \def\yone{3}    % Starting y-coordinate for the first set of boxes
    \def\ytwo{\yone - \vdist}
    \def\ythree{\ytwo - \vdist}
    \def\yfour{\ythree - \vdist}
    \begin{tikzpicture}[
        query/.style={rectangle, draw=yellow!50, fill=yellow!20, thick, minimum height=1.5cm, text width=4.2cm, align=center, font=\Large},
        document/.style={rectangle, draw=green!50, fill=green!20, thick, minimum height=1.5cm, text width=5.9cm, align=center, font=\Large},
        redbox/.style={rectangle, draw=red!50, fill=red!20, thick, minimum height=2cm, text width=2.95cm, align=center, font=\Large},
        bluebox/.style={rectangle, draw=blue!50, fill=blue!20, thick, minimum height=2cm, text width=2.55cm, align=center, font=\Large},
        arrow/.style={->, thick, shorten >=1pt, >=stealth}
    ]
    
    % Labels for each pair (rotated, spaced, and wrapped)
    \node[rotate=90, text=black, text width=2cm, align=center] at (-3.15, \yone) {\Large \textbf{No Lexical Gap}};
    \node[rotate=90, text=black, text width=2.75cm, align=center] at (-3.15, \ytwo) {\Large \textbf{Shallow Gap}};
    \node[rotate=90, text=black, text width=2.75cm, align=center] at (-3.15, \ythree) {\Large \textbf{Inferential Gap}};
    \node[rotate=90, text=black, text width=2.75cm, align=center] at (-3.15, \yfour) {\Large \textbf{Domain Sensitive Query}};
    
    % Query 1
    \node[query] (query1) at (0, \yone) {What is rappelling?};
    \node[document] (doc1) at (6, \yone) {Rappelling is the process of coming down from a mountain...};
    \draw[arrow] (query1) -- (doc1);
    
    % Query 2
    \node[query] (query2) at (0, \ytwo) {Health benefits of antioxidant-rich foods};
    \node[document] (doc2) at (6, \ytwo) {Eating items rich in compounds that fight free radicals can enhance wellness by protecting cells...};
    \draw[arrow] (query2) -- (doc2);
    
    % Query 3
    \node[query] (query3) at (0, \ythree) {Expression of having freedom};
    \node[document] (doc3) at (6, \ythree) {I emptied my car's tank today. Turning 16 could not have been better...};
    \draw[arrow] (query3) -- (doc3);
    
    % Query 4 - New pair with a split document box
    \node[query] (query4) at (0, \yfour) {Expression of having freedom};
    
    % Red box aligned with left edge of other document boxes
    \node[redbox, anchor=west] (doc4a) at (2.9, \yfour) {Go Second Amendment! \#GunOwner};
    
    % Blue box aligned with right edge of other document boxes
    \node[bluebox, anchor=east] (doc4b) at (9, \yfour) {My body, my choice! \#RoeVWade};
    
    % Add labels under the red and blue boxes
    \node at ([yshift=-0.25cm]doc4a.south) {\Large (conservative)};
    \node at ([yshift=-0.25cm]doc4b.south) {\Large  (liberal)};
    
    % Arrow from query to red box
    \draw[arrow] (query4) -- (doc4a);
    
    % Titles or labels
    \node at (0, \yone + 1.25) {\large \textbf{Query}};
    \node at (6, \yone + 1.25) {\large \textbf{Relevant Document}};
    
    \end{tikzpicture}
    }
    \caption{Example of no lexical gap, shallow gap, inferential gap, and domain sensitivity in a retrieval task.}
    \label{fig:gap-example}
\end{figure}
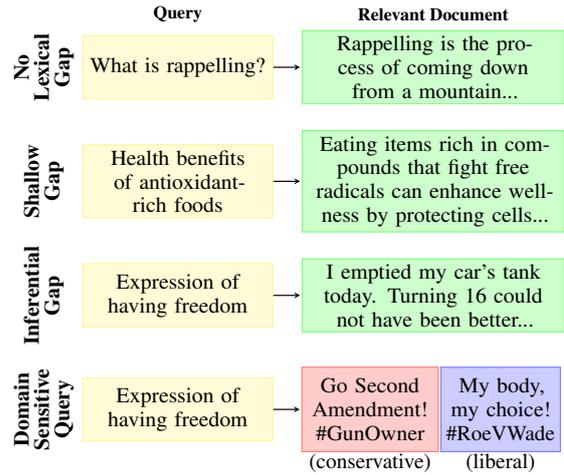

Realizing evidence requires inferring how people express a given trend in text (e.g., when seeking evidence of people having ``freedom'', what kinds of things would they say?). Additionally, understanding the search space itself is crucial for this inference (the evidence for people having ``freedom'', is likely to be different across liberal and conservative communities (Figure~\ref{fig:gap-example})). This motivates the need for a method that can \emph{adapt to the search space}.

Large Language Models (LLMs) have demonstrated remarkable reasoning capabilities in recent years \citep{brown2020languagemodelsfewshotlearners, chen2021evaluatinglargelanguagemodels, rae2022scalinglanguagemodelsmethods}. Since human analysis of large corpora is infeasible, a strong alternative is an LLM that analyzes every single textual document and infers its evidential value. However, doing so would be extremely expensive and time consuming (for example, even a few hundred thousand posts on Reddit would cost thousands of dollars to annotate using OpenAI's GPT-4o model).

In contrast, the Information Retrieval (IR) community has developed light and fast retrieval models capable of searching enormous corpora with impressive accuracy \citep{10.1145/3637870, 10.1145/3486250, zhu2024largelanguagemodelsinformation}. Despite this, these models continue to struggle in cases when the lexical overlap between a query and its relevant documents is low (known as the \emph{lexical gap}) (Figure~\ref{fig:gap-example}) \cite{zhu2024largelanguagemodelsinformation}. While attempts to address the lexical gap have shown promise, they overlook \emph{domain sensitivity}, a problem that occurs when the meaning of the query is sensitive to the search domain (e.g. ``freedom'' for liberals and conservatives), and can only be addressed by adapting to the domain.

This paper aims to bridge the gap between the inefficiency of LLMs and the limited inferential capabilities of IR models, while ensuring adaptability to specific domains. We leverage the strong reasoning capabilities of LLMs to carve out a realization of a trend's evidence within a particular community. We are motivated to use a concept-based method due to its \textbf{interpretability} and \textbf{scalability}. Crucially, our method does not require LLM inference on the whole corpus, operating within a fixed budget of LLM tokens which \textbf{does not depend on the size of the corpus}, making it scalable to large datasets. Our proposed method incrementally allows an LLM to interact with fast retrievers in order to discover the boundaries of the trend being searched \emph{as it is realized in a community} (Figure~\ref{fig:framework-diagram}).

Via this process, our method is able to fit the realization to the community and define a nuanced representation of it. We call this process \textsc{\textbf{ConceptCarve}}, as it is akin to starting with a crude slab of material and carving out a detailed representation of a real-life object. After Concept Carving a trend in human thought patterns, the resulting representation can be used to (1) quickly retrieve evidence from a large dataset (for quantification), and (2) interpret how evidence is realized within a specific community (qualitative analysis). 

Moral Foundations Theory (MFT) is a framework that categorizes the underlying values driving human moral reasoning, and has been thoroughly used by social scientists for two decades \cite{fd3cbcce-5e3f-39a6-b275-9a1d627a0c23, Haidt2007WhenMO, graham2013moral}. These foundations (such as care/harm, fairness/cheating, and loyalty/betrayal) have been shown to manifest differently across groups of people \cite{Khan02042016}. Thus, to reveal the necessity of adaptation to data, we construct a dataset of Reddit posts, partitioned by several communities (e.g. liberal and conservative), and use our approach to find evidence of trends in these moral foundations among various groups. 

We frame `finding evidence' as a reranking task, wherein \textsc{ConceptCarve} supersedes all baselines, achieving a 120.46\% relative improvement in MAP@500 over a dense reranking model, and a 26.03\% relative improvement in MAP@500 over an LLM keyword expansion technique.

To qualitatively analyze the resulting carved representations, we show that they can be used to automatedly detect features that separate two communities for some trend. For example, using our approach, we looked at the trend in \emph{``family members not recognizing desire for freedom''}, among \textbf{liberals} and \textbf{conservatives} and note that evidence of this trend among liberals is realized as discussion of \emph{`personal identity and space'}, while among conservatives, the evidence is realized as \emph{`parental control'} and \emph{`family recognition'}. These analyses demonstrate that \textsc{ConceptCarve} offers significant potential as a tool for both quantifying opinions and capturing their realization across communities. Our contributions can be summarized as: 
\vspace{-2.0em}
\begin{enumerate}
    \setlength{\itemsep}{0pt}
    \setlength{\parskip}{0pt}
    \item Introduce \textsc{ConceptCarve}, a method for dynamically realizing evidence of a trend within a community.
    \item Introduce a dataset which tests a model's ability to deal with (1) inferential gap, and (2) domain sensitivity on the evidence-finding task.\footnote{We release the dataset at \url{https://huggingface.co/datasets/ecaplan/conceptcarve}}
    \item Demonstrate that \textsc{ConceptCarve} outperforms baselines on the dataset.
    \item Use \textsc{ConceptCarve}'s carved representations to analyze how evidence of moral foundations is realized in various communities.
\end{enumerate}

\section{Background and Related Work} IR aims to retrieve relevant documents from large collections based on user queries.

\textbf{Inferential Gap:} The \textbf{lexical gap}, or \emph{vocabulary mismatch}, arises when query and document vocabularies differ. We differentiate between: (1) A \textbf{shallow gap}, resolved by simple rewording to increase overlap, and (2) An \textbf{inferential gap}, requiring complex reasoning and nontrivial inferences (Figure~\ref{fig:gap-example}). Our dataset highlights this inferential gap, which existing datasets do not specifically address. 

Several approaches have been proposed to address the lexical gap problem using LLMs. Generally, the retriever component in Retrieval Augmented Generation (RAG) seeks to improve factuality and memory of a generative agent \cite{gao2024retrievalaugmentedgenerationlargelanguage}. Thus, many works attempt to improve retrieval for an information need \cite{asai2023self, jiang2023activeretrievalaugmentedgeneration, ma-etal-2023-query, pham-etal-2024-whos}, emphasizing generation using top results, not attempting to realize a query within a corpus. 

Several works use LLM embeddings or use LLMs to train smaller retrievers \cite{wang-etal-2024-improving-text, dai2022promptagatorfewshotdenseretrieval, 10.1145/3626772.3657951, yoon-etal-2024-matryoshka, yoon-etal-2024-search}. These works run orthogonal to our own, as our method is agnostic to the retriever, leaving room for any improvements from LLM-based retrievers. Other works use LLMs not as the backbone retriever, but as a tool to reformulate the query \cite{wang2023query2doc, jagerman2023query}. While very cheap, these methods do not interact with the retrieved results, relying solely on an LLM's prediction of relevant results (query expansion).

\textbf{Domain Sensitivity:} We identify the challenge of adapting a domain-sensitive query to the search domain, termed \textbf{ad-hoc domain adaptation} to emphasize that the retriever must adapt to the data \emph{at every query}. Particularly, when there exists an inferential gap, but the inference required to resolve it is highly dependent on the search domain. Figure~\ref{fig:gap-example} demonstrates how the realization of a trend can be extremely sensitive to the dataset being searched.

Various works tackle domain sensitivity via either adapting the retrieval model to the data, or utilizing pseudo-relevance feedback (PRF) \cite{10.1145/383952.383972, 10.1145/1645953.1646259} during search. Parametric methods \cite{saadfalcon2023udapdr, Wang_2022, jiang-etal-2023-boot, zhou-etal-2023-enhancing-generative, wang-etal-2023-effective} adapt the retriever to a specific domain via training. Domain information is also added in other ways, as in \citet{n2022injecting, 10.1162/tacl_a_00530}. Other PRF methods use document embeddings to improve a second retrieval/reranking \cite{Zheng_2020, Wang_2021, gao2022precisezeroshotdenseretrieval, shen2023largelanguagemodelsstrong}, or use an LLM to expand the query with some interactions \cite{jia-etal-2024-mill, weller-etal-2024-generative, chen-etal-2024-analyze, lei-etal-2024-corpus}.

However, these methods tend to: (1) rely on parametric learning; (2) omit recursive refinement of search components; (3) use small initial document sets, limiting their ability to characterize the corpus; or (4) lack interpretability because they are not concept-based, making it difficult to understand the motivation behind the model's results.

Concurrently with our work, \citet{weller2024promptrieverinstructiontrainedretrieversprompted} introduce \emph{Promptriever}, a retrieval model designed to be prompted like a language model. Their method also targets the inferential gap, but does so by parametrically modifying the model through training. In contrast, our approach is concept-based and interpretable, without requiring changes to the retriever itself.

A few works have created abstract, interpretable explanations of text corpora \cite{Lam_2024, NEURIPS2023_7e810b2c, wang-etal-2023-goal}, while others do this by characterizing frames, perspectives, and stances in a social science setting \cite{roy-goldwasser-2023-tale, roy-goldwasser-2020-weakly, reuver-etal-2024-topic, ziems-yang-2021-protect-serve, pujari-goldwasser-2021-understanding, pacheco-etal-2023-interactive}; however, these methods do not use retrievers and are limited to downsampling or smaller datasets when using LLMs.

Most similar to our work is \citet{hoyle-etal-2023-natural}, who represent implicit text explicitly for opinion mining, but must downsample the corpus for LLM use. Likewise, \citet{ravfogel2024descriptionbasedtextsimilarity} search text via abstract descriptions, though they use an LLM for data generation to \emph{train} an encoder. Both methods do not address the issue of domain-sensitivity, which is at the heart of this paper.

\section{Problem Formulation}
In this section, we describe the formal definitions of the tasks, then outline our dataset's construction.

\subsection{Task Definitions}
An \textbf{end-to-end} (E2E) retrieval task considers a set of documents $D$ and a set of queries $Q$. The dataset comes with labeled relevance scores, which can be formulated as a function $\rho: D \times Q \rightarrow \mathbb{R}$, where $\mathbb{R}$ is the set of real numbers, though most datasets either work in binary relevance $\{0,1\}$ or other discrete relevance judgments. The retrieval engine's task is to approximate $\rho$ with its own relevance function, which we denote $\hat{\rho}(d, q)$ for all pairs $(d, q)$. This means that for any pair $(d,q)$, we want to minimize the distance $| \rho(d,q) - \hat{\rho}(d,q) |$.

The \textbf{reranking} task is very similar, except that for each query $q$, a subset of $D$ of size $k$ is selected by the dataset's creators beforehand to be reranked, which we denote $D_q$. Normally, $D_q$ is generated by the top $k$ results of a lightweight, fast retrieval engine on $D$ using $q$. The task is to approximate $\rho$ with a relevance function $\rho_{rerank}: D \times Q \rightarrow \mathbb{R}$, such that $\rho_{rerank}(d, q)$ is similar to $\rho(d,q)$ for all $q$ and $d \in D_q$. In traditional reranking methods, the reranker uses only $D_q$ to rerank documents. However, we believe that even when reranking $D_q$, there is important information garnered from $D$ itself. Subtle observations about the dataset may help inform reranking decisions in a small subset.

We define our task: \textbf{dataset-informed reranking} (DIR), where the task of producing $\rho_{rerank}$ remains the same, but the model has full access to $D$ while reranking $D_q$. To span multiple domains, in our setting we break $D$ into specialized sub-datasets $D_1, D_2, ..., D_n$ and for each, generate a reranking subset $D_{i, q}$ for every query $q \in Q$.

\subsection{Dataset Design}
To fully demonstrate the utility of our framework, we desired the following two features: (1) that the queries require complex inference to relate them to documents (inferential gap) and (2) that the same query manifests differently across different datasets, such as `freedom' in liberal vs. conservative sub-datasets (ad-hoc domain adaptation).

\textbf{Reddit:} We chose Reddit to be our source of social media data because (1) it was diverse in topics and user demographics, (2) it was available via the Cornell ConvoKit project \cite{convokit}, and (3) its segmentation into subreddits enables a natural way of making online communities into sub-datasets. ConvoKit allows access up to 10/2018.

\textbf{Community Definitions:} To obtain sub-datasets, we first defined three pairs of contrasting \uline{communities}: liberal/conservative (political ideology), urban/rural (population density), and religious/secular (spirituality). These were chosen for Reddit availability, distinctiveness, and relevance to social science, for a total of \textbf{six communities}.

\textbf{Community Sub-datasets:} After defining the communities, we collected the top $100,000$ subreddits by size from ConvoKit. We concatenated the subreddit's name with its description, and used an sBERT model's cosine similarity to retrieve the top $10,000$ most similar subreddits for each community. We only kept subreddits which the LLM labeled as `predominantly used by the target community'. We then randomly sampled posts. Table~\ref{tab:community_posts_subreddits} shows the number of posts and subreddits in each community, and Appendix~\ref{app:dataset-details} provides more details, including each community's top subreddits and community overlap. Finally, we had a large dataset $D_c$ for each community $c$.

\begin{table}[ht]
    \small
  \centering
  \begin{tabular}{l c c}
    \hline
    \textbf{Community} & \textbf{\# of Posts (in M)} & \textbf{\# of Subreddits} \\
    \hline
    Conservative          & 44.6  & 268 \\
    Liberal               & 28.2  & 976 \\
    Rural                 & 23.9  & 485 \\
    Urban                 & 19.9  & 1221 \\
    Religious             & 15.4  & 299 \\
    Secular               & 33.6  & 142 \\
    \hline
  \end{tabular}
  \caption{Community, number of posts in the dataset, and number of subreddits sampled to generate it.}
  \label{tab:community_posts_subreddits}
\end{table}

\begin{figure*}[ht!]
    \centering
    % Left side: Original Tree Diagram (now on the left)
    \begin{subfigure}[t]{0.65\textwidth}
    \centering
    \hspace{-3cm}
    \resizebox{\textwidth}{!}{
        \begin{tikzpicture}[
        remember picture,
        node_style/.style={
            rectangle, draw, line width=1.5pt, minimum width=1cm, minimum height=1cm, align=center, font=\Large\ttfamily,
            double copy shadow={shadow xshift=5pt, shadow yshift=-5pt, opacity=.4},
        },
        edge_style/.style={draw, line width=2pt},
        sibling distance=2.5cm,
        level 1/.style={sibling distance=4cm, level distance=1.5cm, edge from parent/.style=edge_style},
        level 2/.style={sibling distance=3.5cm, every child/.style={xshift=0cm}, level distance=1.75cm, edge from parent/.style=edge_style},
        level 3/.style={sibling distance=4cm, level distance=1.75cm, edge from parent/.style=edge_style},
        font=\small,
    ]

    \tikzset{
        red_color/.style={fill={rgb,255:red,255;green,150;blue,150}, text=black},
        green_color/.style={fill={rgb,255:red,170;green,255;blue,170}, text=black},
        blue_color/.style={fill={rgb,255:red,145;green,235;blue,255}, text=black},
        background/.style={rectangle, fill=gray!20, inner sep=0.4cm, rounded corners=5mm},
    }

    \node[node_style, blue_color] (bank) {Bank (+0.5)}
        child {node[node_style, red_color] (cash) {Cash (-0.1)}}
        child {node[node_style, red_color] (money) {Money (-0.2)}}
        child {node[node_style, green_color] (river) {River (+0.23)}
            child {node[node_style, red_color] (fastcurrents) {Fast Currents\\(-0.17)}}
            child {node[node_style, red_color] (polluted) {Polluted\\(-0.12)}}
            child {node[node_style, red_color] (unsafe) {Unsafe\\(-0.12)}}
            child {node[node_style, green_color] (stones) {Stones\\(+0.07)}
                child {node[node_style, red_color] (small) {Small (-0.08)}}
                child {node[node_style, red_color] (rough) {Rough (-0.06)}}
                child {node[node_style, green_color] (big) {Big (+0.01)}}
                child {node[node_style, green_color] (smooth) {Smooth (+0.02)}}}
            child {node[node_style, green_color] (shallow) {Shallow\\(+0.08)}}
            child {node[node_style, green_color] (clearwater) {Clear Water\\(+0.09)}}};

    % Background rectangle with coordinates
    \begin{scope}[on background layer]
        \node[background, fit=(clearwater), name=smallbox, inner sep=15pt] {};
    \end{scope}
    % Coordinates for corners of the small gray box
    \coordinate (smallboxtr) at (smallbox.north east);
    \coordinate (smallboxbr) at (smallbox.south east);

    \end{tikzpicture}
    }
    \end{subfigure}
    % \hspace{1cm}
    % \vspace{-0.5cm}
    % \hspace{1cm}
    % Right side: Zoomed-in Clear Water with additional nodes (now on the right)
    \begin{subfigure}[t]{0.15\textwidth}
        \begin{tikzpicture}[
            remember picture,
            node_style/.style={
                rectangle, draw, minimum width=1.5cm, minimum height=0.5cm, align=center, font=\scriptsize\ttfamily,
            },
            light_node_style/.style={
                rectangle, draw, minimum width=0.5cm, minimum height=0.75cm, text width=2.5cm, align=center, font=\scriptsize, inner sep=1pt
            },
        ]

        \def\top{-2}
        \def\bot{1}
        \def\left{0}
        \def\right{4}
        
        % Gray box background with coordinates for corners
        \coordinate (bigboxbl) at (\left, \bot);
        \coordinate (bigboxtl) at (\left, \top);
        \coordinate (bigboxbr) at (\right, \bot);
        \coordinate (bigboxtr) at (\right, \top);
        \begin{scope}
            \fill[gray!20, rounded corners=0.5cm] (bigboxbl) rectangle (bigboxtr);
        \end{scope}

        \clip (current bounding box.south west) rectangle (current bounding box.north east);

        % Define colors
        \definecolor{green_color}{RGB}{170, 255, 170}
        \colorlet{light_green_color}{green_color!40!white}

        % Node styles
        \tikzset{
            green_color_node/.style={fill=green_color, text=black},
            light_green_color_node/.style={fill=light_green_color, text=black},
        }

        % Node positions
        \node[node_style, green_color_node] (clearwater) at ($(1.4, 2.5) + (\left, \top)$) {Clear Water (+0.09)};

        % Offsets
        \def\dx{0.25}
        \def\dy{-0.715}
        \def\dxone{0.6}
        \def\dyone{-0.575}

        % Additional nodes
        \node[light_node_style, light_green_color_node] (node1) at ($(clearwater) + (\dxone,\dyone)$) {``You can see everything---it's safe.''};

        \node[light_node_style, light_green_color_node] (node2) at ($(node1) + (\dx,\dy)$) {``This water’s super clean.''};

        \node[light_node_style, light_green_color_node] (node3) at ($(node2) + (\dx,\dy)$) {``Nothing scary is hiding under there.''};

        \node[light_node_style, light_green_color_node] (node2) at ($(node2)$) {``This water’s super clean.''};

        \node[light_node_style, light_green_color_node] (node1) at ($(node1)$) {``You can see everything---it's safe.''};

        \node[node_style, green_color_node] (clearwater) at ($(clearwater)$) {Clear Water (+0.09)};

        \end{tikzpicture}
    \end{subfigure}

    % Draw lines connecting the corners
    \begin{tikzpicture}[remember picture, overlay]
        \draw[-, ultra thick, gray!20] ($(bigboxtl) + (0.15cm,0.15cm)$) -- (2.3, 1.15);
        \draw[-, ultra thick, gray!20] ($(bigboxbl) + (0.18cm,-0.15cm)$) -- (2.3, 2.05);
    \end{tikzpicture}

    \begin{tikzpicture}[remember picture, overlay]
    \node at (current page.center) [yshift=12cm] { % Adjust yshift as needed
        \parbox[b]{5.75cm}{ % Adjust width as needed
            \centering
            \small
            \textbf{Intent:} ``give descriptions of the river bank of a river that has big, smooth stones, and is safe for swimming''
        }
    };
\end{tikzpicture}

    \caption{\small(Left): Concept tree: promoted (green) and demoted (red) concepts. Note that this tree sets the root to the ambiguous `Bank'; our Characterizer would have used the full intent. (Right): Concept of ``Clear Water'', displayed with its set of groundings.}
    \label{fig:bank-example}
\end{figure*}
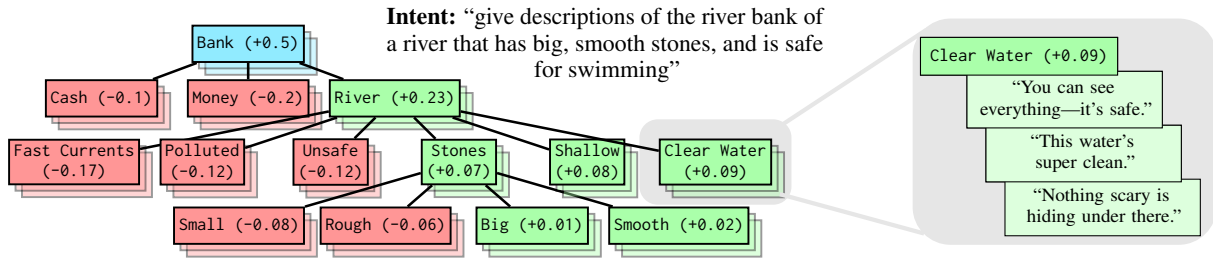

\textbf{Trends:} Independently of the communities and our framework, we systematically generated complex, domain-sensitive trends to be our queries. To ensure they were \textbf{domain-sensitive}, we chose each query to be a trend based on a moral foundation, since we know that moral foundations manifest differently across communities \cite{Khan02042016}. First, we created a list of 6 base trends, each of the form ``Increase in belief that people feel X'', where ``X'' is the positive end of the moral foundation (caring, fairness, loyalty etc.) 

Since we desired an \textbf{inferential gap}, we did not want simple trends whose evidence could be determined by discussion topic. Thus, to induce the queries to be more complex, we defined several \uline{complexity dimensions}: time (how opinions change over time), relationships (influence of social connections), evidence (whether opinions are based on personal or objective sources), emotions (impact on emotional states), agency (who is acting or affected), and scope (whether changes occur at an individual or societal level). Then, we asked an LLM to qualify each trend with 5 distinct combinations of 2-3 complexity dimensions. The result was 30 complex, domain-sensitive trends, which we used as our set of queries $Q$. The full list of trends can be found in Appendix~\ref{app:dataset-details}.

\textbf{Reranking Set:} 
To construct the reranking subset $D_{c,q}$ for each query $q$ and community $c$, a mix of relevant and non-relevant posts was needed. Since the trends are so specific, finding relevant posts was nontrivial. We used the ColBERT retriever \cite{Khattab_2020} to form a reranking set for each pair $c, q$. For each query, $k=2000$ posts were retrieved from $D_c$ to form the subset $D_{c,q}$. ColBERT was chosen for its speed and semantic retrieval capabilities, as simpler lexical searches likely wouldn't yield sufficient evidence. The choice of retrieval engine does not affect the final task since the goal is to rerank the results.

\textbf{Labeling} To label the relevance of each post in $D_{c,q}$, we had an LLM label each $d \in D_{c,q}$ by asking if $d$ is evidence of $q$ (prompt in Appendix~\ref{app:prompts}, no mention of community $c$). This 0/1 label was used for evaluating reranking. The final dataset consists of: (1) 30 independent, complex, domain-sensitive trends, (2) 6 community sub-datasets $D_{rural}, D_{urban},$... (3) 180 sets of 2000 posts, and each post labeled as evidence/not evidence (one set for each trend/community pair).

\textbf{Human Validation}
Over 12 hours of human annotations show 68\% agreement with LLM labels and 70\% inter-annotator agreement (see Appendix~\ref{app:dataset-details}). While our goal is to align with LLM judgments rather than humans', these results minimally validate the LLM's judgments for this task.

\section{Concept Carve}
Here, we define terms used in the \textsc{ConceptCarve} framework, followed by the framework itself.

\subsection{Definitions}

\textbf{Intent:} An \emph{intent} is the user's goal in retrieval, expressed as text. In our setting, the intent is to ``find evidence'' for a trend in human behavior, but it could be any goal. For example, ``Find evidence that teens are using vaping products in public schools.'' Alternatively, it could be ``get ideas,'' like ``Give me ideas on how to decorate my room using minimalism.'' Figure~\ref{fig:bank-example} shows a complex intent.

\noindent
\textbf{Grounding:} A \emph{grounding} is a single string which can be used as a query with a standard retriever. Figure~\ref{fig:bank-example} (right) depicts several examples.

\noindent
\textbf{Concept:} A \emph{concept} is an abstract idea, represented by a set of groundings (as seen in Figure~\ref{fig:bank-example}). For convenience, concepts can be named; though only groundings are ever used. A concept acts as a bridge: concrete enough for retrieval (using groundings), yet compact enough for LLM reasoning. 

\noindent
\textbf{Concept Tree:} A \emph{concept tree} is a tree of weighted concepts. Positively-weighted concepts are \emph{promoted} and negatively-weighted concepts are \emph{demoted}. By carefully adding promoted and demoted concepts, the tree can carve out a complex intent (Figure~\ref{fig:bank-example}). For example, it allows for promoting a broad idea while downplaying certain aspects, in addition to allowing for iterative refinement by interacting with the data.

\subsection{Framework}

\captionsetup[figure]{labelfont={small}, textfont={small}}
\begin{figure*}[h]
    \centering
    \includegraphics[width=\linewidth]{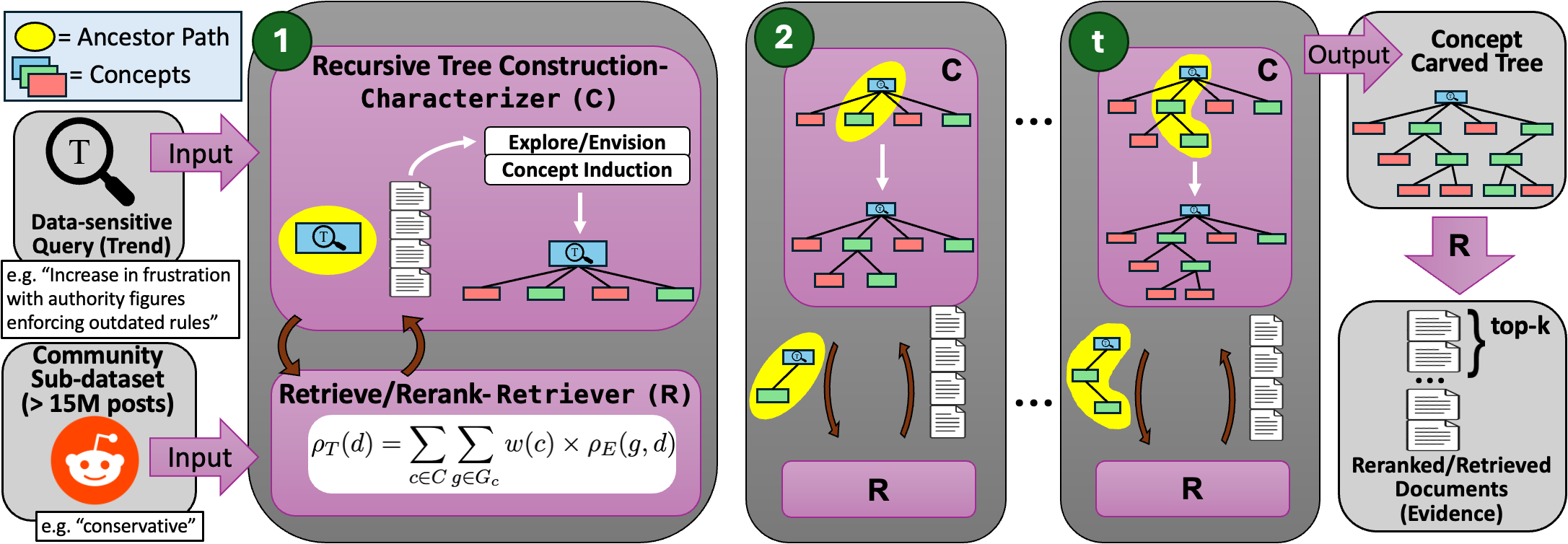}
    \caption{\small\textsc{ConceptCarve}: a concept tree is recursively constructed in steps $1\ldots t$ by alternating between Characterizer (generates new concepts from intermediate docs) and Retriever (retrieves docs using intermediate trees). Output tree represents the realization of the input trend within the input community and can be used for evidence retrieval or analyzed directly.}
    \label{fig:framework-diagram}
\end{figure*}

The framework consists of the main module, Characterizer, and its submodule, Retriever. The Characterizer interactively grows a concept tree, repeatedly using the Retriever to get intermediate results. Figure~\ref{fig:framework-diagram} gives an overview of the framework.

\subsubsection{Retriever}

The \emph{Retriever} utilizes an off-the-shelf retrieval engine, $E$ and a concept tree $T$ (which may be partially built) to either rerank or retrieve from a set of documents $D$. In the Characterizer section, we explain how the Retriever's operations are used repeatedly to grow a tree. While most retrieval systems use a query to calculate a document's relevance score, the Retriever uses a concept tree. $T$ can embody some (potentially) complicated idea, and so we use its concepts' groundings and the off-the-shelf retriever $E$ to calculate the score.

To compute the relevance score for a document $d$ to the tree $T$, we use all of its concepts' groundings: Let $C$ be the set of concepts in $T$, and $G_c$ be the set of groundings for concept $c$. Let $w(c)$ denote the weight of concept $c$ in $T$. Finally, let $\rho_E(g, d)$ denote the relevance score assigned by $E$ to $d$ with grounding $g$. Then $T$'s assigned relevance score to $d$ is given by equation~\ref{eq:rep-relevance}.
\vspace{-0.5em}
\begin{equation}
\rho_T(d) = \sum_{c \in C} \sum_{g \in G_c} w(c) \times \rho_E(g, d)
\label{eq:rep-relevance}
\end{equation}
\vspace{-1em}

We simply use the off-the-shelf engine to find the relevance of the document for each grounding of each concept, and add them up, weighting them via the respective concept's weight. We do this for every document in a small set $D_q$ to perform reranking, as shown in equation~\ref{eq:rep-rerank}. We can also do this scoring for every document in the whole dataset $D$, and rank all documents to perform end-to-end retrieval. When this is the case, the retriever returns a set of top $k$ documents according to the scoring function, as shown in equation~\ref{eq:rep-end-to-end-retrieve}. \vspace{-0.35em}
\begin{equation}
\operatorname{rerank}(T, D_q) = \{ (d, \rho_T(d)) \mid d \in D_q \}
\label{eq:rep-rerank}
\end{equation}
\vspace{-1em}
\begin{equation}
\fontsize{10.375pt}{11pt}\selectfont
\operatorname{retrieve}(T, D, k) = \text{top-}k \{ (d, \rho_T(d)) \mid d \in D \}
\label{eq:rep-end-to-end-retrieve}
\end{equation}

Two things should be noted: (1) demoted concept weights are negative, and hence relevance to demoted concepts reduces the relevance of a document, and (2) the concept tree structure of $T$ is not taken into account when calculating a document's relevance to $T$. That is, the simplicity of the relevance scoring relies heavily on the high quality of the concepts, groundings, and weights of $T$.

\subsubsection{Characterizer}
The \emph{Characterizer} grows a concept tree for some intent over some dataset. At a high level, it does this by judiciously using an LLM to inspect retrieved documents, forming concepts, and reasoning over which concepts should be promoted and which should be demoted. Importantly, the Characterizer is the only component which requires reasoning, and therefore is the only component which requires an LLM. The input to the Characterizer is an intent $i$, a dataset $D$, and a $k$ indicating the size of each intermediate retrieval. Its output is a carved concept tree $T$ representing $i$ as it manifests in $D$.

The algorithm begins with a root concept (single-node tree), which, when retrieved, should provide a starting point for the Characterizer. In our implementation, we create the root by making a concept whose grounding is just $i$, though any initialization is possible (as in Figure~\ref{fig:bank-example}).

To grow out a concept's children, the Characterizer performs three high-level operations: (1) \textbf{ancestor path retrieval:} retrieve using an intermediate concept tree, (2) \textbf{envision/explore:} cluster the retrieved documents, creating groups of documents that support/refute the intent, and (3) \textbf{concept induction:} extract properties from each group and use them to generate groundings for new concepts (the new children). To grow the whole tree, these operations are performed recursively, starting at the root. We now explain each operation in detail.

\begin{table*}[!ht]
    \small
    \centering
    \begin{tabular}{l|ccc|ccc|ccc}
        \hline
        \textbf{Reranking System} & \multicolumn{3}{c|}{@10} & \multicolumn{3}{c|}{@100} & \multicolumn{3}{c}{@500} \\
        \cline{2-10}
        & P & R & MAP & P & R & MAP & P & R & MAP \\
        \hline
        BM25 & 13.20 & 0.70 & 0.30 & 12.90 & 6.10 & 1.10 & 12.70 & 27.50 & 3.80 \\
        ColBERT & 26.10 & 1.30 & 0.60 & 21.00 & 9.20 & 2.50 & 16.70 & 34.80 & 7.10 \\
        ANCE & 23.70 & 1.30 & 0.60 & 18.70 & 8.70 & 2.20 & 16.00 & 33.40 & 6.50 \\
        RepLLaMA & 14.11 & 0.53 & 0.23 & 13.42 & 5.04 & 0.94 & 15.05 & 29.84 & 4.49 \\
        Query2Doc + ColBERT & 37.28 & 2.20 & 1.33 & 26.57 & 13.42 & 4.82 & 19.59 & 42.43 & 11.37 \\
        MultiQuery + ColBERT & 25.20 & 1.33 & 0.71 & 19.89 & 9.49 & 2.60 & 16.42 & 35.49 & 7.08 \\
        \textsc{Envision Only} & 38.00 & 2.10 & 1.20 & 28.20 & 14.40 & 5.10 & 20.70 & 46.00 & 12.50 \\
        \textsc{ConceptCarve} (depth 1) & \underline{40.11} & \underline{2.39} & \underline{1.46} & \underline{29.83} & \underline{15.75} & \underline{5.80} & \underline{21.44} & \underline{48.86} & \underline{13.81} \\
        \textsc{ConceptCarve} (depth 2) & \textbf{41.56} & \textbf{2.40} & \textbf{1.49} & \textbf{30.70} & \textbf{16.38} & \textbf{6.10} & \textbf{21.78} & \textbf{49.71} & \textbf{14.33} \\
        \hline
    \end{tabular}
    \caption{Performance on the DIR task. The best is \textbf{bolded}, and the second best \underline{underlined}.}
    \label{tab:dataset-informed-results}
\end{table*}

\textbf{Ancestor Path Retrieval:}
Given a concept $c$, we wish to inspect which documents it contributes to the whole tree's retrieval. Since $c$ may depend on other concepts in the tree for its meaning, we isolate its entire ancestor path—the subtree containing all nodes from the root to $c$. Then, using the ancestor path as a concept tree in itself, we apply the Retriever's $\operatorname{retrieve}$ operation to get the top $k$ documents from the dataset $D$ that are most relevant to the path. We call this set $D_{ret}$.

\textbf{Envision/Explore:}
The \textbf{explore} operation aims to inspect $D_{ret}$ and find useful ideas within it, while the \textbf{envision} operation aims to inspect $D_{ret}$ and expand the search space by introducing new ideas. First, we cluster $D_{ret}$ using BERTopic \cite{grootendorst2022bertopic}. In both operations, the top $m$ clusters (each with $n$ centroid documents) are presented to an LLM. In \textbf{explore}, the LLM identifies clusters that support or refute the intent. For \textbf{envision}, an LLM generates centroids that the LLM deems \emph{should} support the intent but are missing from the clusters. The result of both operations is a set of clusters that either support or refute the intent.

\textbf{Concept Induction:}
The final step is to convert these clusters into concepts, a process we call \textbf{concept induction}. To create a concept, we need a set of groundings. For supporting clusters, we provide an LLM with the cluster's centroids and the intent $i$, asking it to generate properties explaining why the documents support $i$. The same process is done for refuting clusters, explaining why they do not support $i$. The LLM then synthesizes these properties into artificial documents, which serve as the groundings for the new concept. For convenience, the LLM also names the concepts, though the names are not used in retrieval.

\textbf{Weighting:}
The weighting scheme chosen gives lesser weights to children than parents, ensures equality among siblings, and normalizes the weights. Intuitively, this means that a subconcept can only partially counteract its superconcept. Details are in Appendix~\ref{app:weighting}.

\textbf{Reranking:} Finally, since we wish to use the final concept tree $T$ for reranking a set of documents $D_q$, we can simply call $\operatorname{rerank}(T, D_q)$.

\section{DIR Evaluation}

\begin{figure}[h]
    \centering
    \includegraphics[width=\linewidth]{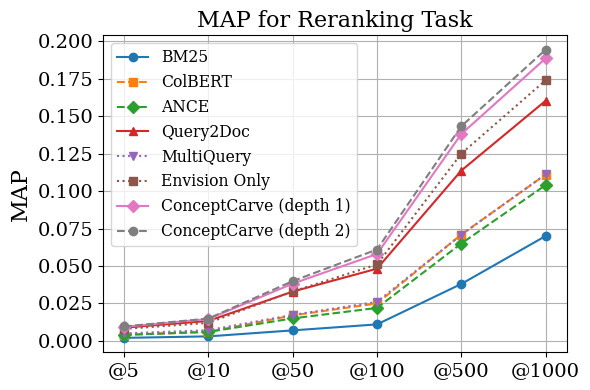}
    \caption{MAP@k on DIR task.}
    \label{fig:map-plot}
\end{figure}

This section details Dataset-informed Reranking (DIR), performed on the constructed dataset.

\textsc{ConceptCarve} allows various tree configurations. We can limit the promoted (PBF), demoted (DBF), and envisioned branching factors (EBF), as well as tree depth. After pilot tests, we set PBF, EBF, and DBF to 5, with a maximum depth of 2, since concept weights diminished beyond that. For reranking, we only used promoted concept nodes, as the benefit of demoted concepts in reranking was not seen (see end-to-end retrieval as to why).

Our baselines were as follows. \textbf{BM25:} \cite{inproceedings} a lexical retrieval model. \textbf{ANCE:} \cite{xiong2020approximate} a dense, exact search bi-encoder model. \textbf{ColBERT:} \cite{Khattab_2020} a late interaction retriever. \textbf{Query2Doc + ColBERT:} \cite{wang2023query2doc} an LLM expands the query into a hypothetical document, then reranks with ColBERT. \textbf{MultiQuery + ColBERT:} \cite{langchain_multiqueryretriever} an LLM rewrites the query three times, then reranks using ColBERT on all three. \textbf{RepLLaMA:} \cite{ma2023finetuningllamamultistagetext} a dense retriever which uses LLaMA-2-7B as its backbone. \textbf{\textsc{EnvisionOnly}:} a version of \textsc{ConceptCarve} where concepts come only from the \textbf{envision} operation (what the LLM sees as missing from intermediate results). Reproducibility details in Appendix~\ref{app:reproducibility}.

\textbf{DIR Results:}
Table~\ref{tab:dataset-informed-results} and Figure~\ref{fig:map-plot} show that \textsc{ConceptCarve} outperforms all baseline reranking models. Models using an LLM, including the \textsc{EnvisionOnly} and Query2Doc, significantly outperform dense and lexical models, highlighting the LLM's ability to address the inferential gap. Both \textsc{ConceptCarve} models, especially at depth 2, surpass \textsc{EnvisionOnly} and Query2Doc, demonstrating the benefit of interacting with the data during tree construction. This shows the utility of the \textbf{explore} operation and the ability of the dataset to test ad hoc domain adaptation. Finally, depth 2 slightly outperforms depth 1, indicating that exploring more concepts improves trend realization.

\textbf{Generality and Dataset Diversity:}
While our evaluation uses a single data source (Reddit), the dataset spans over 3,000 subreddits within a wide range of domains, including politics, religion, hobbies, employment, education, and online culture. This results in over 165 million posts for tree construction and 360,000 for reranking. The diversity in topic, style, and community norms offers a broad and challenging testbed for evidence realization. Notably, \textsc{ConceptCarve} operates without any training or tuning, using only an off-the-shelf ColBERT retriever and an LLM, yet performs very well across this diverse data. We argue that this setting provides strong evidence of the framework’s capacity to generalize across communities and topics in the task of realizing evidence of human behavior and opinion.

\section{Discussion and Analysis}
We detail E2E retrieval analysis, a qualitative analysis of the trees, and the costs of \textsc{ConceptCarve}.

\textbf{E2E Retrieval:}
Demoting concepts in pilot experiments didn't improve reranking. This may be because the reranking set was made from ColBERT's top $2000$ results, which were already aligned with the trend, lacking noise to be removed. To test this, we performed E2E retrieval using concept trees with, and without demoted concepts. In reranking, the document set is fixed, but here we measure how many `evidence' examples are retrieved, using only P@k as a metric, labeling on the fly. Since such labeling is expensive, we used 24 trend-community pairs (48K posts), and analysis was limited to testing demoted concepts.

\begin{table}
    \small
    \setlength{\tabcolsep}{3pt}
    \centering
  \begin{tabular}{lcccccc}
    \hline
    \textbf{Retriever} & P@5 & P@10 & P@50 & P@100 & P@500 & P@1K \\
    \hline
    ColBERT & 27.8 & 25.4 & 22.5 & 20.9 & 16.7 & 14.9 \\
    \textsc{CC} ($+$) & 30.8 & \textbf{34.2} & 29.8 & 25.8 & 19.8 & 17.9\\
    \textsc{CC} ($+-$) & \textbf{34.2} & 32.9 & \textbf{30.7} & \textbf{26.9} & \textbf{20.4} & \textbf{18.0}\\
    \hline
    \end{tabular}
    \caption{Performance on the end-to-end retrieval task, with ($+-$) and without ($+$) demoted concepts.}
    \label{tab:retrieval-ablation}
\end{table}

\textbf{E2E Results:} Table~\ref{tab:retrieval-ablation} shows the E2E results, ablating the demoted concepts. The results show that including demoted concepts slightly improves the precision of the retrieved results. We hypothesize that the improvement is small since the retrieval process is very sensitive to the weighting difference in promoted and demoted concepts. Despite this, the results support our hypothesis that demoted concepts reduce the relevance score of irrelevant posts when retrieving from the full dataset.

\textbf{Concept Tree Qualitative Analysis:}
We show that concept-carved trees provide interpretable realizations of trends across communities. To compare communities, we construct concept trees for the same trend within opposing communities (e.g., rural vs. urban). During construction, each concept is grounded via concept induction, where an LLM identifies key properties that make evidence ``supporting.'' These are then analyzed to extract polarity—differences in evidence priority between communities (e.g., ``mentions/does not mention mental health''). These polarities are visualized to highlight how communities realize evidence differently.

\begin{figure}[!ht]
    \centering
    \includegraphics[width=\linewidth]{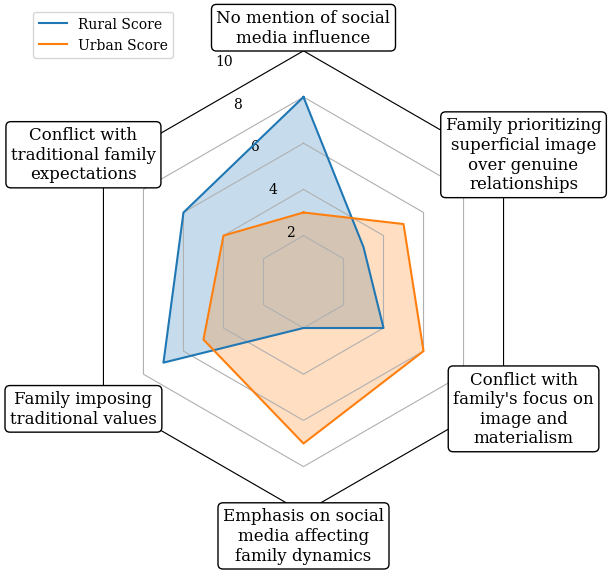}
    \caption{Trend: ``Increase in frustration with family members who seem to prioritize personal ambitions over traditional family values.''}
    \label{fig:spider3}
\end{figure}

In Figure~\ref{fig:spider3}, we compare the concept trees for the \emph{rural} and \emph{urban} communities. Using the trees' properties, an LLM identifies polarities and scores their usefulness in identifying evidence for the trend in each community. The spider plot shows that, for the urban community, social media effects strongly indicate evidence. In contrast, rural evidence emphasizes conflict over traditional family expectations, while urban evidence focuses on conflict related to a family's image. Plots for other trends/communities are in Appendix~\ref{app:more-qual-analysis}. This analysis reveals both \emph{how} \textsc{ConceptCarve} retrieves evidence, and qualitatively demonstrates that the trees represent each community's realization.

\textbf{Cost Analysis:}  
\textsc{ConceptCarve}'s cost includes characterization and retrieval. The Characterizer's dominant cost is the number of LLM calls, measured asymptotically in input tokens. With $B$ as the branching factor, $m$ as the clusters shown during envision/explore, and $n$ as the centroid documents per cluster, the Characterizer's cost is $O(B^2n + Bmn)$. In our implementation, this equates to $\sim$20,000 tokens per tree, \textbf{independent of dataset size} and number of reranked/retrieved documents $k$. Thus, while our method incurs a higher initial query cost, it scales efficiently to massive datasets. Retrieval cost, measured in retriever calls, is $O(C \times \gamma)$, where $\gamma$ is the number of groundings per concept and $C$ is the total number of concepts. Appendix~\ref{app:cost-analysis} provides detailed derivations and examines accuracy trade-offs.

\section{Conclusion}
We introduced \textsc{ConceptCarve}, a retrieval framework that combines traditional retrievers with LLM-guided concept construction to adapt to specific communities. It addresses two major challenges in evidence retrieval: the inferential gap and ad-hoc domain adaptation. Unlike LLM-based query expansion or embedding methods, \textsc{ConceptCarve} iteratively builds a concept tree, leading to stronger performance and greater interpretability.

Our experiments show that \textsc{ConceptCarve} outperforms both traditional and LLM-augmented baselines, despite using no training or fine-tuning. Its effectiveness across a large, diverse dataset highlights its potential for retrieving and analyzing how communities express complex opinions and behaviors.

Future work includes refining concept weighting schemes and exploring how trends evolve over time or align with real-world events. We hope this framework inspires further research at the intersection of retrieval, reasoning, and human-centered analysis.

\section*{Limitations} The Reddit data used in our experiments does not represent the full spectrum of online discussions, limiting the generalizability of our results to other platforms or domains. Additionally, LLMs were used extensively for data annotation, which introduces potential biases inherent in these models. While we manually validated some LLM labels, the overall quality and fairness of the labels may still be affected by the limitations of the LLMs themselves.

\section*{Ethics Statement}
LLMs were used extensively in this work, and we acknowledge their potential for bias due to the nature of their training data. All products of LLMs, including complex frameworks, should be critically evaluated and not taken at face value when real-world consequences are involved. To mitigate risks, we applied human oversight whenever possible. All datasets used were either publicly available or collected with proper consent, ensuring data privacy. We are committed to ethical AI use, fairness, and transparency throughout this project.

\bibliography{custom}

\begin{thebibliography}{66}
\providecommand{\natexlab}[1]{#1}

\bibitem[{Asai et~al.(2023)Asai, Wu, Wang, Sil, and Hajishirzi}]{asai2023self}
Akari Asai, Zeqiu Wu, Yizhong Wang, Avirup Sil, and Hannaneh Hajishirzi. 2023.
\newblock Self-rag: Learning to retrieve, generate, and critique through self-reflection.
\newblock \emph{arXiv preprint arXiv:2310.11511}.

\bibitem[{Brown et~al.(2020)Brown, Mann, Ryder, Subbiah, Kaplan, Dhariwal, Neelakantan, Shyam, Sastry, Askell, Agarwal, Herbert-Voss, Krueger, Henighan, Child, Ramesh, Ziegler, Wu, Winter, Hesse, Chen, Sigler, Litwin, Gray, Chess, Clark, Berner, McCandlish, Radford, Sutskever, and Amodei}]{brown2020languagemodelsfewshotlearners}
Tom~B. Brown, Benjamin Mann, Nick Ryder, Melanie Subbiah, Jared Kaplan, Prafulla Dhariwal, Arvind Neelakantan, Pranav Shyam, Girish Sastry, Amanda Askell, Sandhini Agarwal, Ariel Herbert-Voss, Gretchen Krueger, Tom Henighan, Rewon Child, Aditya Ramesh, Daniel~M. Ziegler, Jeffrey Wu, Clemens Winter, Christopher Hesse, Mark Chen, Eric Sigler, Mateusz Litwin, Scott Gray, Benjamin Chess, Jack Clark, Christopher Berner, Sam McCandlish, Alec Radford, Ilya Sutskever, and Dario Amodei. 2020.
\newblock \href {https://arxiv.org/abs/2005.14165} {Language models are few-shot learners}.
\newblock \emph{Preprint}, arXiv:2005.14165.

\bibitem[{Chang et~al.(2020)Chang, Chiam, Fu, Wang, Zhang, and Danescu-Niculescu-Mizil}]{convokit}
Jonathan~P. Chang, Caleb Chiam, Liye Fu, Andrew Wang, Justine Zhang, and Cristian Danescu-Niculescu-Mizil. 2020.
\newblock Convokit: A toolkit for the analysis of conversations.
\newblock In \emph{Proceedings of SIGDIAL}.

\bibitem[{Chase(2022)}]{langchain_multiqueryretriever}
Harrison Chase. 2022.
\newblock \href {https://api.python.langchain.com/en/latest/retrievers/langchain.retrievers.multi_query.MultiQueryRetriever.html} {Langchain: Multiqueryretriever method documentation}.
\newblock Accessed: 2024-12-11.
\newblock Version 1.2.0.

\bibitem[{Chen et~al.(2021)Chen, Tworek, Jun, Yuan, de~Oliveira~Pinto, Kaplan, Edwards, Burda, Joseph, Brockman, Ray, Puri, Krueger, Petrov, Khlaaf, Sastry, Mishkin, Chan, Gray, Ryder, Pavlov, Power, Kaiser, Bavarian, Winter, Tillet, Such, Cummings, Plappert, Chantzis, Barnes, Herbert-Voss, Guss, Nichol, Paino, Tezak, Tang, Babuschkin, Balaji, Jain, Saunders, Hesse, Carr, Leike, Achiam, Misra, Morikawa, Radford, Knight, Brundage, Murati, Mayer, Welinder, McGrew, Amodei, McCandlish, Sutskever, and Zaremba}]{chen2021evaluatinglargelanguagemodels}
Mark Chen, Jerry Tworek, Heewoo Jun, Qiming Yuan, Henrique~Ponde de~Oliveira~Pinto, Jared Kaplan, Harri Edwards, Yuri Burda, Nicholas Joseph, Greg Brockman, Alex Ray, Raul Puri, Gretchen Krueger, Michael Petrov, Heidy Khlaaf, Girish Sastry, Pamela Mishkin, Brooke Chan, Scott Gray, Nick Ryder, Mikhail Pavlov, Alethea Power, Lukasz Kaiser, Mohammad Bavarian, Clemens Winter, Philippe Tillet, Felipe~Petroski Such, Dave Cummings, Matthias Plappert, Fotios Chantzis, Elizabeth Barnes, Ariel Herbert-Voss, William~Hebgen Guss, Alex Nichol, Alex Paino, Nikolas Tezak, Jie Tang, Igor Babuschkin, Suchir Balaji, Shantanu Jain, William Saunders, Christopher Hesse, Andrew~N. Carr, Jan Leike, Josh Achiam, Vedant Misra, Evan Morikawa, Alec Radford, Matthew Knight, Miles Brundage, Mira Murati, Katie Mayer, Peter Welinder, Bob McGrew, Dario Amodei, Sam McCandlish, Ilya Sutskever, and Wojciech Zaremba. 2021.
\newblock \href {https://arxiv.org/abs/2107.03374} {Evaluating large language models trained on code}.
\newblock \emph{Preprint}, arXiv:2107.03374.

\bibitem[{Chen et~al.(2024)Chen, Chen, He, Wen, and Sun}]{chen-etal-2024-analyze}
Xinran Chen, Xuanang Chen, Ben He, Tengfei Wen, and Le~Sun. 2024.
\newblock \href {https://doi.org/10.18653/v1/2024.findings-acl.708} {Analyze, generate and refine: Query expansion with {LLM}s for zero-shot open-domain {QA}}.
\newblock In \emph{Findings of the Association for Computational Linguistics: ACL 2024}, pages 11908--11922, Bangkok, Thailand. Association for Computational Linguistics.

\bibitem[{Dai et~al.(2022)Dai, Zhao, Ma, Luan, Ni, Lu, Bakalov, Guu, Hall, and Chang}]{dai2022promptagatorfewshotdenseretrieval}
Zhuyun Dai, Vincent~Y. Zhao, Ji~Ma, Yi~Luan, Jianmo Ni, Jing Lu, Anton Bakalov, Kelvin Guu, Keith~B. Hall, and Ming-Wei Chang. 2022.
\newblock \href {https://arxiv.org/abs/2209.11755} {Promptagator: Few-shot dense retrieval from 8 examples}.
\newblock \emph{Preprint}, arXiv:2209.11755.

\bibitem[{Gao et~al.(2022)Gao, Ma, Lin, and Callan}]{gao2022precisezeroshotdenseretrieval}
Luyu Gao, Xueguang Ma, Jimmy Lin, and Jamie Callan. 2022.
\newblock \href {https://arxiv.org/abs/2212.10496} {Precise zero-shot dense retrieval without relevance labels}.
\newblock \emph{Preprint}, arXiv:2212.10496.

\bibitem[{Gao et~al.(2024)Gao, Xiong, Gao, Jia, Pan, Bi, Dai, Sun, Wang, and Wang}]{gao2024retrievalaugmentedgenerationlargelanguage}
Yunfan Gao, Yun Xiong, Xinyu Gao, Kangxiang Jia, Jinliu Pan, Yuxi Bi, Yi~Dai, Jiawei Sun, Meng Wang, and Haofen Wang. 2024.
\newblock \href {https://arxiv.org/abs/2312.10997} {Retrieval-augmented generation for large language models: A survey}.
\newblock \emph{Preprint}, arXiv:2312.10997.

\bibitem[{Graham et~al.(2013)Graham, Haidt, Koleva, Motyl, Iyer, Wojcik, and Ditto}]{graham2013moral}
Jesse Graham, Jonathan Haidt, Sena Koleva, Matt Motyl, Ravi Iyer, Sean~P Wojcik, and Peter~H Ditto. 2013.
\newblock Moral foundations theory: The pragmatic validity of moral pluralism.
\newblock In \emph{Advances in experimental social psychology}, volume~47, pages 55--130. Elsevier.

\bibitem[{Grootendorst(2022)}]{grootendorst2022bertopic}
Maarten Grootendorst. 2022.
\newblock Bertopic: Neural topic modeling with a class-based tf-idf procedure.
\newblock \emph{arXiv preprint arXiv:2203.05794}.

\bibitem[{Guo et~al.(2022)Guo, Cai, Fan, Sun, Zhang, and Cheng}]{10.1145/3486250}
Jiafeng Guo, Yinqiong Cai, Yixing Fan, Fei Sun, Ruqing Zhang, and Xueqi Cheng. 2022.
\newblock \href {https://doi.org/10.1145/3486250} {Semantic models for the first-stage retrieval: A comprehensive review}.
\newblock \emph{ACM Trans. Inf. Syst.}, 40(4).

\bibitem[{Haidt and Graham(2007)}]{Haidt2007WhenMO}
Jonathan Haidt and Jesse Graham. 2007.
\newblock \href {https://api.semanticscholar.org/CorpusID:6824095} {When morality opposes justice: Conservatives have moral intuitions that liberals may not recognize}.
\newblock \emph{Social Justice Research}, 20:98--116.

\bibitem[{Haidt and Joseph(2004)}]{fd3cbcce-5e3f-39a6-b275-9a1d627a0c23}
Jonathan Haidt and Craig Joseph. 2004.
\newblock \href {http://www.jstor.org/stable/20027945} {Intuitive ethics: How innately prepared intuitions generate culturally variable virtues}.
\newblock \emph{Daedalus}, 133(4):55--66.

\bibitem[{Hashimoto et~al.(2024)Hashimoto, Takazawa, and Sasahara}]{hashimoto2024meat}
Moena Hashimoto, Yotaro Takazawa, and Kazutoshi Sasahara. 2024.
\newblock Are meat alternatives a moral concern? a comparison of english and japanese tweets.
\newblock \emph{Humanities and Social Sciences Communications}, 11(1):1--10.

\bibitem[{He et~al.(2024)He, Wang, Zhao, and Yang}]{He_Wang_Zhao_Yang_2024}
Bo~He, Lei Wang, Ruoyu Zhao, and Yongfen Yang. 2024.
\newblock \href {https://doi.org/10.54097/2v77z645} {Analysis of online public opinion texts based on topic mining and sentiment analysis}.
\newblock \emph{Mathematical Modeling and Algorithm Application}, 2(2):23–28.

\bibitem[{Hoyle et~al.(2023)Hoyle, Sarkar, Goel, and Resnik}]{hoyle-etal-2023-natural}
Alexander Hoyle, Rupak Sarkar, Pranav Goel, and Philip Resnik. 2023.
\newblock \href {https://doi.org/10.18653/v1/2023.emnlp-main.815} {Natural language decompositions of implicit content enable better text representations}.
\newblock In \emph{Proceedings of the 2023 Conference on Empirical Methods in Natural Language Processing}, pages 13188--13214, Singapore. Association for Computational Linguistics.

\bibitem[{Jagerman et~al.(2023)Jagerman, Zhuang, Qin, Wang, and Bendersky}]{jagerman2023query}
Rolf Jagerman, Honglei Zhuang, Zhen Qin, Xuanhui Wang, and Michael Bendersky. 2023.
\newblock Query expansion by prompting large language models.
\newblock \emph{arXiv preprint arXiv:2305.03653}.

\bibitem[{Jia et~al.(2024)Jia, Liu, Zhao, Li, Hao, Wang, and Yin}]{jia-etal-2024-mill}
Pengyue Jia, Yiding Liu, Xiangyu Zhao, Xiaopeng Li, Changying Hao, Shuaiqiang Wang, and Dawei Yin. 2024.
\newblock \href {https://doi.org/10.18653/v1/2024.naacl-long.138} {{MILL}: Mutual verification with large language models for zero-shot query expansion}.
\newblock In \emph{Proceedings of the 2024 Conference of the North American Chapter of the Association for Computational Linguistics: Human Language Technologies (Volume 1: Long Papers)}, pages 2498--2518, Mexico City, Mexico. Association for Computational Linguistics.

\bibitem[{Jiang et~al.(2023{\natexlab{a}})Jiang, Xu, Drummond, and Cohn}]{jiang-etal-2023-boot}
Fan Jiang, Qiongkai Xu, Tom Drummond, and Trevor Cohn. 2023{\natexlab{a}}.
\newblock \href {https://doi.org/10.18653/v1/2023.findings-emnlp.65} {Boot and switch: Alternating distillation for zero-shot dense retrieval}.
\newblock In \emph{Findings of the Association for Computational Linguistics: EMNLP 2023}, pages 912--931, Singapore. Association for Computational Linguistics.

\bibitem[{Jiang et~al.(2023{\natexlab{b}})Jiang, Xu, Gao, Sun, Liu, Dwivedi-Yu, Yang, Callan, and Neubig}]{jiang2023activeretrievalaugmentedgeneration}
Zhengbao Jiang, Frank~F. Xu, Luyu Gao, Zhiqing Sun, Qian Liu, Jane Dwivedi-Yu, Yiming Yang, Jamie Callan, and Graham Neubig. 2023{\natexlab{b}}.
\newblock \href {https://arxiv.org/abs/2305.06983} {Active retrieval augmented generation}.
\newblock \emph{Preprint}, arXiv:2305.06983.

\bibitem[{Kang et~al.(2023)Kang, Park, Jo, and Bak}]{kang-etal-2023-values}
Dongjun Kang, Joonsuk Park, Yohan Jo, and JinYeong Bak. 2023.
\newblock \href {https://doi.org/10.18653/v1/2023.emnlp-main.961} {From values to opinions: Predicting human behaviors and stances using value-injected large language models}.
\newblock In \emph{Proceedings of the 2023 Conference on Empirical Methods in Natural Language Processing}, pages 15539--15559, Singapore. Association for Computational Linguistics.

\bibitem[{Khan and Stagnaro(2016)}]{Khan02042016}
Saera~R. Khan and Michael~Nick Stagnaro. 2016.
\newblock \href {https://doi.org/10.1080/10508422.2015.1007997} {The influence of multiple group identities on moral foundations}.
\newblock \emph{Ethics \& Behavior}, 26(3):194--214.

\bibitem[{Khattab and Zaharia(2020)}]{Khattab_2020}
Omar Khattab and Matei Zaharia. 2020.
\newblock \href {https://doi.org/10.1145/3397271.3401075} {Colbert: Efficient and effective passage search via contextualized late interaction over bert}.
\newblock In \emph{Proceedings of the 43rd International ACM SIGIR Conference on Research and Development in Information Retrieval}, SIGIR ’20, page 39–48. ACM.

\bibitem[{Lam et~al.(2024)Lam, Teoh, Landay, Heer, and Bernstein}]{Lam_2024}
Michelle~S. Lam, Janice Teoh, James~A. Landay, Jeffrey Heer, and Michael~S. Bernstein. 2024.
\newblock \href {https://doi.org/10.1145/3613904.3642830} {Concept induction: Analyzing unstructured text with high-level concepts using lloom}.
\newblock In \emph{Proceedings of the CHI Conference on Human Factors in Computing Systems}, CHI ’24, page 1–28. ACM.

\bibitem[{Lavrenko and Croft(2001)}]{10.1145/383952.383972}
Victor Lavrenko and W.~Bruce Croft. 2001.
\newblock \href {https://doi.org/10.1145/383952.383972} {Relevance based language models}.
\newblock In \emph{Proceedings of the 24th Annual International ACM SIGIR Conference on Research and Development in Information Retrieval}, SIGIR '01, page 120–127, New York, NY, USA. Association for Computing Machinery.

\bibitem[{Lei et~al.(2024)Lei, Cao, Zhou, Shen, and Yates}]{lei-etal-2024-corpus}
Yibin Lei, Yu~Cao, Tianyi Zhou, Tao Shen, and Andrew Yates. 2024.
\newblock \href {https://aclanthology.org/2024.eacl-short.34} {Corpus-steered query expansion with large language models}.
\newblock In \emph{Proceedings of the 18th Conference of the European Chapter of the Association for Computational Linguistics (Volume 2: Short Papers)}, pages 393--401, St. Julian{'}s, Malta. Association for Computational Linguistics.

\bibitem[{Lv and Zhai(2009)}]{10.1145/1645953.1646259}
Yuanhua Lv and ChengXiang Zhai. 2009.
\newblock \href {https://doi.org/10.1145/1645953.1646259} {A comparative study of methods for estimating query language models with pseudo feedback}.
\newblock In \emph{Proceedings of the 18th ACM Conference on Information and Knowledge Management}, CIKM '09, page 1895–1898, New York, NY, USA. Association for Computing Machinery.

\bibitem[{Ma et~al.(2023{\natexlab{a}})Ma, Gong, He, Zhao, and Duan}]{ma-etal-2023-query}
Xinbei Ma, Yeyun Gong, Pengcheng He, Hai Zhao, and Nan Duan. 2023{\natexlab{a}}.
\newblock \href {https://doi.org/10.18653/v1/2023.emnlp-main.322} {Query rewriting in retrieval-augmented large language models}.
\newblock In \emph{Proceedings of the 2023 Conference on Empirical Methods in Natural Language Processing}, pages 5303--5315, Singapore. Association for Computational Linguistics.

\bibitem[{Ma et~al.(2023{\natexlab{b}})Ma, Wang, Yang, Wei, and Lin}]{ma2023finetuningllamamultistagetext}
Xueguang Ma, Liang Wang, Nan Yang, Furu Wei, and Jimmy Lin. 2023{\natexlab{b}}.
\newblock \href {https://arxiv.org/abs/2310.08319} {Fine-tuning llama for multi-stage text retrieval}.
\newblock \emph{Preprint}, arXiv:2310.08319.

\bibitem[{Ma et~al.(2024)Ma, Wang, Yang, Wei, and Lin}]{10.1145/3626772.3657951}
Xueguang Ma, Liang Wang, Nan Yang, Furu Wei, and Jimmy Lin. 2024.
\newblock \href {https://doi.org/10.1145/3626772.3657951} {Fine-tuning llama for multi-stage text retrieval}.
\newblock In \emph{Proceedings of the 47th International ACM SIGIR Conference on Research and Development in Information Retrieval}, SIGIR '24, page 2421–2425, New York, NY, USA. Association for Computing Machinery.

\bibitem[{Pacheco et~al.(2023)Pacheco, Islam, Ungar, Yin, and Goldwasser}]{pacheco-etal-2023-interactive}
Maria~Leonor Pacheco, Tunazzina Islam, Lyle Ungar, Ming Yin, and Dan Goldwasser. 2023.
\newblock \href {https://doi.org/10.18653/v1/2023.findings-acl.313} {Interactive concept learning for uncovering latent themes in large text collections}.
\newblock In \emph{Findings of the Association for Computational Linguistics: ACL 2023}, pages 5059--5080, Toronto, Canada. Association for Computational Linguistics.

\bibitem[{Paulissen and Wendt(2023)}]{paulissen-wendt-2023-lauri}
Spencer Paulissen and Caroline Wendt. 2023.
\newblock \href {https://doi.org/10.18653/v1/2023.semeval-1.27} {Lauri ingman at {S}em{E}val-2023 task 4: A chain classifier for identifying human values behind arguments}.
\newblock In \emph{Proceedings of the 17th International Workshop on Semantic Evaluation (SemEval-2023)}, pages 193--198, Toronto, Canada. Association for Computational Linguistics.

\bibitem[{Pham et~al.(2024)Pham, Ngo, Luu, and Nguyen}]{pham-etal-2024-whos}
Quang~Hieu Pham, Hoang Ngo, Anh~Tuan Luu, and Dat~Quoc Nguyen. 2024.
\newblock \href {https://doi.org/10.18653/v1/2024.findings-emnlp.593} {Who{'}s who: Large language models meet knowledge conflicts in practice}.
\newblock In \emph{Findings of the Association for Computational Linguistics: EMNLP 2024}, pages 10142--10151, Miami, Florida, USA. Association for Computational Linguistics.

\bibitem[{Pila{\v{r}} et~al.(2021)Pila{\v{r}}, Kvasni{\v{c}}kov{\'a}~Stanislavsk{\'a}, and Kvasni{\v{c}}ka}]{pilavr2021healthy}
Ladislav Pila{\v{r}}, Lucie Kvasni{\v{c}}kov{\'a}~Stanislavsk{\'a}, and Roman Kvasni{\v{c}}ka. 2021.
\newblock Healthy food on the twitter social network: Vegan, homemade, and organic food.
\newblock \emph{International journal of environmental research and public health}, 18(7):3815.

\bibitem[{Pujari and Goldwasser(2021)}]{pujari-goldwasser-2021-understanding}
Rajkumar Pujari and Dan Goldwasser. 2021.
\newblock \href {https://doi.org/10.18653/v1/2021.emnlp-main.102} {Understanding politics via contextualized discourse processing}.
\newblock In \emph{Proceedings of the 2021 Conference on Empirical Methods in Natural Language Processing}, pages 1353--1367, Online and Punta Cana, Dominican Republic. Association for Computational Linguistics.

\bibitem[{Qorib et~al.(2023)Qorib, Oladunni, Denis, Ososanya, and Cotae}]{qorib2023covid}
Miftahul Qorib, Timothy Oladunni, Max Denis, Esther Ososanya, and Paul Cotae. 2023.
\newblock Covid-19 vaccine hesitancy: Text mining, sentiment analysis and machine learning on covid-19 vaccination twitter dataset.
\newblock \emph{Expert Systems with Applications}, 212:118715.

\bibitem[{Rae et~al.(2022)Rae, Borgeaud, Cai, Millican, Hoffmann, Song, Aslanides, Henderson, Ring, Young, Rutherford, Hennigan, Menick, Cassirer, Powell, van~den Driessche, Hendricks, Rauh, Huang, Glaese, Welbl, Dathathri, Huang, Uesato, Mellor, Higgins, Creswell, McAleese, Wu, Elsen, Jayakumar, Buchatskaya, Budden, Sutherland, Simonyan, Paganini, Sifre, Martens, Li, Kuncoro, Nematzadeh, Gribovskaya, Donato, Lazaridou, Mensch, Lespiau, Tsimpoukelli, Grigorev, Fritz, Sottiaux, Pajarskas, Pohlen, Gong, Toyama, de~Masson~d'Autume, Li, Terzi, Mikulik, Babuschkin, Clark, de~Las~Casas, Guy, Jones, Bradbury, Johnson, Hechtman, Weidinger, Gabriel, Isaac, Lockhart, Osindero, Rimell, Dyer, Vinyals, Ayoub, Stanway, Bennett, Hassabis, Kavukcuoglu, and Irving}]{rae2022scalinglanguagemodelsmethods}
Jack~W. Rae, Sebastian Borgeaud, Trevor Cai, Katie Millican, Jordan Hoffmann, Francis Song, John Aslanides, Sarah Henderson, Roman Ring, Susannah Young, Eliza Rutherford, Tom Hennigan, Jacob Menick, Albin Cassirer, Richard Powell, George van~den Driessche, Lisa~Anne Hendricks, Maribeth Rauh, Po-Sen Huang, Amelia Glaese, Johannes Welbl, Sumanth Dathathri, Saffron Huang, Jonathan Uesato, John Mellor, Irina Higgins, Antonia Creswell, Nat McAleese, Amy Wu, Erich Elsen, Siddhant Jayakumar, Elena Buchatskaya, David Budden, Esme Sutherland, Karen Simonyan, Michela Paganini, Laurent Sifre, Lena Martens, Xiang~Lorraine Li, Adhiguna Kuncoro, Aida Nematzadeh, Elena Gribovskaya, Domenic Donato, Angeliki Lazaridou, Arthur Mensch, Jean-Baptiste Lespiau, Maria Tsimpoukelli, Nikolai Grigorev, Doug Fritz, Thibault Sottiaux, Mantas Pajarskas, Toby Pohlen, Zhitao Gong, Daniel Toyama, Cyprien de~Masson~d'Autume, Yujia Li, Tayfun Terzi, Vladimir Mikulik, Igor Babuschkin, Aidan Clark, Diego de~Las~Casas, Aurelia Guy, Chris Jones,
  James Bradbury, Matthew Johnson, Blake Hechtman, Laura Weidinger, Iason Gabriel, William Isaac, Ed~Lockhart, Simon Osindero, Laura Rimell, Chris Dyer, Oriol Vinyals, Kareem Ayoub, Jeff Stanway, Lorrayne Bennett, Demis Hassabis, Koray Kavukcuoglu, and Geoffrey Irving. 2022.
\newblock \href {https://arxiv.org/abs/2112.11446} {Scaling language models: Methods, analysis \& insights from training gopher}.
\newblock \emph{Preprint}, arXiv:2112.11446.

\bibitem[{Ravfogel et~al.(2024)Ravfogel, Pyatkin, Cohen, Manevich, and Goldberg}]{ravfogel2024descriptionbasedtextsimilarity}
Shauli Ravfogel, Valentina Pyatkin, Amir~DN Cohen, Avshalom Manevich, and Yoav Goldberg. 2024.
\newblock \href {https://arxiv.org/abs/2305.12517} {Description-based text similarity}.
\newblock \emph{Preprint}, arXiv:2305.12517.

\bibitem[{Reuver et~al.(2024)Reuver, Polimeno, Fokkens, and Lopes}]{reuver-etal-2024-topic}
Myrthe Reuver, Alessandra Polimeno, Antske Fokkens, and Ana~Isabel Lopes. 2024.
\newblock \href {https://aclanthology.org/2024.cpss-1.8} {Topic-specific social science theory in stance detection: a proposal and interdisciplinary pilot study on sustainability initiatives}.
\newblock In \emph{Proceedings of the 4th Workshop on Computational Linguistics for the Political and Social Sciences: Long and short papers}, pages 101--111, Vienna, Austria. Association for Computational Linguistics.

\bibitem[{Robertson et~al.(1994)Robertson, Walker, Jones, Hancock-Beaulieu, and Gatford}]{inproceedings}
Stephen Robertson, Steve Walker, Susan Jones, Micheline Hancock-Beaulieu, and Mike Gatford. 1994.
\newblock Okapi at trec-3.
\newblock pages 0--.

\bibitem[{Roy and Goldwasser(2020)}]{roy-goldwasser-2020-weakly}
Shamik Roy and Dan Goldwasser. 2020.
\newblock \href {https://doi.org/10.18653/v1/2020.emnlp-main.620} {Weakly supervised learning of nuanced frames for analyzing polarization in news media}.
\newblock In \emph{Proceedings of the 2020 Conference on Empirical Methods in Natural Language Processing (EMNLP)}, pages 7698--7716, Online. Association for Computational Linguistics.

\bibitem[{Roy and Goldwasser(2023)}]{roy-goldwasser-2023-tale}
Shamik Roy and Dan Goldwasser. 2023.
\newblock \href {https://doi.org/10.18653/v1/2023.findings-emnlp.701} {{``}a tale of two movements{'}: Identifying and comparing perspectives in {\#}{B}lack{L}ives{M}atter and {\#}{B}lue{L}ives{M}atter movements-related tweets using weakly supervised graph-based structured prediction}.
\newblock In \emph{Findings of the Association for Computational Linguistics: EMNLP 2023}, pages 10437--10467, Singapore. Association for Computational Linguistics.

\bibitem[{Saad-Falcon et~al.(2023)Saad-Falcon, Khattab, Santhanam, Florian, Franz, Roukos, Sil, Sultan, and Potts}]{saadfalcon2023udapdr}
Jon Saad-Falcon, Omar Khattab, Keshav Santhanam, Radu Florian, Martin Franz, Salim Roukos, Avirup Sil, Md~Arafat Sultan, and Christopher Potts. 2023.
\newblock \href {https://arxiv.org/abs/2303.00807} {Udapdr: Unsupervised domain adaptation via llm prompting and distillation of rerankers}.
\newblock \emph{Preprint}, arXiv:2303.00807.

\bibitem[{Shen et~al.(2023)Shen, Long, Geng, Tao, Zhou, and Jiang}]{shen2023largelanguagemodelsstrong}
Tao Shen, Guodong Long, Xiubo Geng, Chongyang Tao, Tianyi Zhou, and Daxin Jiang. 2023.
\newblock \href {https://arxiv.org/abs/2304.14233} {Large language models are strong zero-shot retriever}.
\newblock \emph{Preprint}, arXiv:2304.14233.

\bibitem[{Siriwardhana et~al.(2023)Siriwardhana, Weerasekera, Wen, Kaluarachchi, Rana, and Nanayakkara}]{10.1162/tacl_a_00530}
Shamane Siriwardhana, Rivindu Weerasekera, Elliott Wen, Tharindu Kaluarachchi, Rajib Rana, and Suranga Nanayakkara. 2023.
\newblock \href {https://doi.org/10.1162/tacl_a_00530} {{Improving the Domain Adaptation of Retrieval Augmented Generation (RAG) Models for Open Domain Question Answering}}.
\newblock \emph{Transactions of the Association for Computational Linguistics}, 11:1--17.

\bibitem[{Thakur et~al.(2022)Thakur, Reimers, and Lin}]{n2022injecting}
Nandan Thakur, Nils Reimers, and Jimmy Lin. 2022.
\newblock \href {https://arxiv.org/abs/2205.11498} {Injecting domain adaptation with learning-to-hash for effective and efficient zero-shot dense retrieval}.
\newblock \emph{Preprint}, arXiv:2205.11498.

\bibitem[{Thakur et~al.(2021)Thakur, Reimers, Rücklé, Srivastava, and Gurevych}]{thakur2021beirheterogenousbenchmarkzeroshot}
Nandan Thakur, Nils Reimers, Andreas Rücklé, Abhishek Srivastava, and Iryna Gurevych. 2021.
\newblock \href {https://arxiv.org/abs/2104.08663} {Beir: A heterogenous benchmark for zero-shot evaluation of information retrieval models}.
\newblock \emph{Preprint}, arXiv:2104.08663.

\bibitem[{Wang et~al.(2022)Wang, Thakur, Reimers, and Gurevych}]{Wang_2022}
Kexin Wang, Nandan Thakur, Nils Reimers, and Iryna Gurevych. 2022.
\newblock \href {https://doi.org/10.18653/v1/2022.naacl-main.168} {Gpl: Generative pseudo labeling for unsupervised domain adaptation of dense retrieval}.
\newblock In \emph{Proceedings of the 2022 Conference of the North American Chapter of the Association for Computational Linguistics: Human Language Technologies}. Association for Computational Linguistics.

\bibitem[{Wang et~al.(2024)Wang, Yang, Huang, Yang, Majumder, and Wei}]{wang-etal-2024-improving-text}
Liang Wang, Nan Yang, Xiaolong Huang, Linjun Yang, Rangan Majumder, and Furu Wei. 2024.
\newblock \href {https://doi.org/10.18653/v1/2024.acl-long.642} {Improving text embeddings with large language models}.
\newblock In \emph{Proceedings of the 62nd Annual Meeting of the Association for Computational Linguistics (Volume 1: Long Papers)}, pages 11897--11916, Bangkok, Thailand. Association for Computational Linguistics.

\bibitem[{Wang et~al.(2023{\natexlab{a}})Wang, Yang, and Wei}]{wang2023query2doc}
Liang Wang, Nan Yang, and Furu Wei. 2023{\natexlab{a}}.
\newblock \href {https://arxiv.org/abs/2303.07678} {Query2doc: Query expansion with large language models}.
\newblock \emph{Preprint}, arXiv:2303.07678.

\bibitem[{Wang et~al.(2023{\natexlab{b}})Wang, MacAvaney, Macdonald, and Ounis}]{wang-etal-2023-effective}
Xiao Wang, Sean MacAvaney, Craig Macdonald, and Iadh Ounis. 2023{\natexlab{b}}.
\newblock \href {https://doi.org/10.18653/v1/2023.acl-long.710} {Effective contrastive weighting for dense query expansion}.
\newblock In \emph{Proceedings of the 61st Annual Meeting of the Association for Computational Linguistics (Volume 1: Long Papers)}, pages 12688--12704, Toronto, Canada. Association for Computational Linguistics.

\bibitem[{Wang et~al.(2021)Wang, Macdonald, Tonellotto, and Ounis}]{Wang_2021}
Xiao Wang, Craig Macdonald, Nicola Tonellotto, and Iadh Ounis. 2021.
\newblock \href {https://doi.org/10.1145/3471158.3472250} {Pseudo-relevance feedback for multiple representation dense retrieval}.
\newblock In \emph{Proceedings of the 2021 ACM SIGIR International Conference on Theory of Information Retrieval}, ICTIR ’21. ACM.

\bibitem[{Wang et~al.(2023{\natexlab{c}})Wang, Shang, and Zhong}]{wang-etal-2023-goal}
Zihan Wang, Jingbo Shang, and Ruiqi Zhong. 2023{\natexlab{c}}.
\newblock \href {https://doi.org/10.18653/v1/2023.emnlp-main.657} {Goal-driven explainable clustering via language descriptions}.
\newblock In \emph{Proceedings of the 2023 Conference on Empirical Methods in Natural Language Processing}, pages 10626--10649, Singapore. Association for Computational Linguistics.

\bibitem[{Weller et~al.(2024{\natexlab{a}})Weller, Durme, Lawrie, Paranjape, Zhang, and Hessel}]{weller2024promptrieverinstructiontrainedretrieversprompted}
Orion Weller, Benjamin~Van Durme, Dawn Lawrie, Ashwin Paranjape, Yuhao Zhang, and Jack Hessel. 2024{\natexlab{a}}.
\newblock \href {https://arxiv.org/abs/2409.11136} {Promptriever: Instruction-trained retrievers can be prompted like language models}.
\newblock \emph{Preprint}, arXiv:2409.11136.

\bibitem[{Weller et~al.(2024{\natexlab{b}})Weller, Lo, Wadden, Lawrie, Van~Durme, Cohan, and Soldaini}]{weller-etal-2024-generative}
Orion Weller, Kyle Lo, David Wadden, Dawn Lawrie, Benjamin Van~Durme, Arman Cohan, and Luca Soldaini. 2024{\natexlab{b}}.
\newblock \href {https://aclanthology.org/2024.findings-eacl.134} {When do generative query and document expansions fail? a comprehensive study across methods, retrievers, and datasets}.
\newblock In \emph{Findings of the Association for Computational Linguistics: EACL 2024}, pages 1987--2003, St. Julian{'}s, Malta. Association for Computational Linguistics.

\bibitem[{Xiong et~al.(2020)Xiong, Xiong, Li, Tang, Liu, Bennett, Ahmed, and Overwijk}]{xiong2020approximate}
Lee Xiong, Chenyan Xiong, Ye~Li, Kwok-Fung Tang, Jialin Liu, Paul Bennett, Junaid Ahmed, and Arnold Overwijk. 2020.
\newblock \href {https://arxiv.org/abs/2007.00808} {Approximate nearest neighbor negative contrastive learning for dense text retrieval}.
\newblock \emph{Preprint}, arXiv:2007.00808.

\bibitem[{Yoon et~al.(2024{\natexlab{a}})Yoon, Chen, Arik, and Pfister}]{yoon-etal-2024-search}
Jinsung Yoon, Yanfei Chen, Sercan Arik, and Tomas Pfister. 2024{\natexlab{a}}.
\newblock \href {https://doi.org/10.18653/v1/2024.acl-long.661} {Search-adaptor: Embedding customization for information retrieval}.
\newblock In \emph{Proceedings of the 62nd Annual Meeting of the Association for Computational Linguistics (Volume 1: Long Papers)}, pages 12230--12247, Bangkok, Thailand. Association for Computational Linguistics.

\bibitem[{Yoon et~al.(2024{\natexlab{b}})Yoon, Sinha, Arik, and Pfister}]{yoon-etal-2024-matryoshka}
Jinsung Yoon, Rajarishi Sinha, Sercan~O Arik, and Tomas Pfister. 2024{\natexlab{b}}.
\newblock \href {https://doi.org/10.18653/v1/2024.emnlp-main.576} {Matryoshka-adaptor: Unsupervised and supervised tuning for smaller embedding dimensions}.
\newblock In \emph{Proceedings of the 2024 Conference on Empirical Methods in Natural Language Processing}, pages 10318--10336, Miami, Florida, USA. Association for Computational Linguistics.

\bibitem[{Zhang et~al.(2023)Zhang, Wang, and Liu}]{zhang2023moral}
Weiyu Zhang, Rong Wang, and Haodong Liu. 2023.
\newblock Moral expressions, sources, and frames: Examining covid-19 vaccination posts by facebook public pages.
\newblock \emph{Computers in Human Behavior}, 138:107479.

\bibitem[{Zhao et~al.(2024)Zhao, Liu, Ren, and Wen}]{10.1145/3637870}
Wayne~Xin Zhao, Jing Liu, Ruiyang Ren, and Ji-Rong Wen. 2024.
\newblock \href {https://doi.org/10.1145/3637870} {Dense text retrieval based on pretrained language models: A survey}.
\newblock \emph{ACM Trans. Inf. Syst.}, 42(4).

\bibitem[{Zheng et~al.(2020)Zheng, Hui, He, Han, Sun, and Yates}]{Zheng_2020}
Zhi Zheng, Kai Hui, Ben He, Xianpei Han, Le~Sun, and Andrew Yates. 2020.
\newblock \href {https://doi.org/10.18653/v1/2020.findings-emnlp.424} {Bert-qe: Contextualized query expansion for document re-ranking}.
\newblock In \emph{Findings of the Association for Computational Linguistics: EMNLP 2020}. Association for Computational Linguistics.

\bibitem[{Zhong et~al.(2023)Zhong, Zhang, Li, Ahn, Klein, and Steinhardt}]{NEURIPS2023_7e810b2c}
Ruiqi Zhong, Peter Zhang, Steve Li, Jinwoo Ahn, Dan Klein, and Jacob Steinhardt. 2023.
\newblock \href {https://proceedings.neurips.cc/paper_files/paper/2023/file/7e810b2c75d69be186cadd2fe3febeab-Paper-Conference.pdf} {Goal driven discovery of distributional differences via language descriptions}.
\newblock In \emph{Advances in Neural Information Processing Systems}, volume~36, pages 40204--40237. Curran Associates, Inc.

\bibitem[{Zhou et~al.(2023)Zhou, Dou, and Wen}]{zhou-etal-2023-enhancing-generative}
Yujia Zhou, Zhicheng Dou, and Ji-Rong Wen. 2023.
\newblock \href {https://doi.org/10.18653/v1/2023.emnlp-main.768} {Enhancing generative retrieval with reinforcement learning from relevance feedback}.
\newblock In \emph{Proceedings of the 2023 Conference on Empirical Methods in Natural Language Processing}, pages 12481--12490, Singapore. Association for Computational Linguistics.

\bibitem[{Zhu et~al.(2024)Zhu, Yuan, Wang, Liu, Liu, Deng, Chen, Liu, Dou, and Wen}]{zhu2024largelanguagemodelsinformation}
Yutao Zhu, Huaying Yuan, Shuting Wang, Jiongnan Liu, Wenhan Liu, Chenlong Deng, Haonan Chen, Zheng Liu, Zhicheng Dou, and Ji-Rong Wen. 2024.
\newblock \href {https://arxiv.org/abs/2308.07107} {Large language models for information retrieval: A survey}.
\newblock \emph{Preprint}, arXiv:2308.07107.

\bibitem[{Ziems and Yang(2021)}]{ziems-yang-2021-protect-serve}
Caleb Ziems and Diyi Yang. 2021.
\newblock \href {https://doi.org/10.18653/v1/2021.findings-emnlp.82} {To protect and to serve? analyzing entity-centric framing of police violence}.
\newblock In \emph{Findings of the Association for Computational Linguistics: EMNLP 2021}, pages 957--976, Punta Cana, Dominican Republic. Association for Computational Linguistics.

\end{thebibliography}

\appendix
\section{Dataset Details}
\label{app:dataset-details}

\begin{table*}[h!]
    \small
    \centering
    \begin{tabular}{l c l | l c l}
        \hline
        \textbf{Community} & \textbf{\# of Posts (K)} & \textbf{Top Subreddits} & \textbf{Community} & \textbf{\# of Posts (K)} & \textbf{Top Subreddits} \\
        \hline
        Conservative & 1000 & personalfinance     & Liberal      & 200  & Anarchism             \\
                     & 1000 & The\_Donald         &              & 200  & SandersForPresident   \\
                     & 1000 & Frugal              &              & 200  & Political\_Revolution  \\
                     & 1000 & Libertarian         &              & 200  & WayOfTheBern          \\
                     & 1000 & Conservative        &              & 200  & Socialism             \\
                     & 1000 & MGTOW               &              & 200  & AskALiberal           \\
                     & 1000 & ar15                &              & 200  & Feminism              \\
                     & 1000 & Firearms            &              & 200  & Futurology            \\
                     & 1000 & MensRights          &              & 200  & LateStageCapitalism   \\
                     & 1000 & Patriots            &              & 200  & ChapoTrapHouse        \\
        \hline
        Religious    & 5000 & Christianity        & Secular      & 3000 & atheism               \\
                     & 1704 & Catholicism         &              & 3000 & Futurology            \\
                     & 1514 & islam               &              & 3000 & science               \\
                     & 973  & Psychonaut          &              & 3000 & exmormon              \\
                     & 889  & Buddhism            &              & 3000 & askscience            \\
                     & 776  & DebateAnAtheist     &              & 2876 & DebateReligion        \\
                     & 773  & Judaism             &              & 2741 & space                 \\
                     & 633  & Meditation          &              & 1658 & exjw                  \\
                     & 528  & TrueChristian       &              & 1513 & philosophy            \\
                     & 486  & latterdaysaints     &              & 1251 & Anarchism             \\
        \hline
        Rural        & 780  & motorcycles         & Urban        & 75   & CitiesSkylines        \\
                     & 780  & woodworking         &              & 75   & nyc                   \\
                     & 780  & environment         &              & 75   & baltimore             \\
                     & 780  & DIY                 &              & 75   & toronto               \\
                     & 780  & ireland             &              & 75   & shanghai              \\
                     & 780  & gardening           &              & 75   & Tokyo                 \\
                     & 780  & Firearms            &              & 75   & BravoRealHousewives   \\
                     & 780  & Fishing             &              & 75   & vancouver             \\
                     & 780  & dogs                &              & 75   & cincinnati            \\
                     & 780  & NASCAR              &              & 75   & kansascity            \\
        \hline
    \end{tabular}
    \caption{Top 10 subreddits in each community by number of posts sampled from that subreddit.}
    \label{tab:community_top_subreddits}
\end{table*}

\begin{table*}[h!]
    \newcommand{\parboxsize}{1cm} % Define the parbox size variable
    \footnotesize
    \centering
    \begin{tabular}{ll} % Adjust width as necessary
        \hline
        \textbf{MF} & \textbf{Trend (``Increase in...'')} \\
        \hline
        \multirow{6}{*}{\rotatebox{90}{\parbox{\parboxsize}{\textbf{Care/\\Harm}}}} &
        individuals expressing guilt over not caring for their community, while \\
        & \quad acknowledging external influences. \\
        & people feeling mixed gratitude and frustration over care from close friends. \\
        & disappointment with younger generations over care shown to older people, despite \\
        & \quad reports of improvement. \\
        & people saying they feel cared for by others but express uneasiness about it. \\
        & belief that it's acceptable to cause harm to certain groups based on historical actions. \\
        \hline
        \multirow{6}{*}{\rotatebox{90}{\parbox{\parboxsize}{\textbf{Fairness/\\Cheating}}}} &
        the belief that fairness is more prevalent locally than nationally. \\
        & perception of unfairness toward older adults, even if not personally experienced. \\
        & frustration toward claims of unfairness based on personal stories over broader evidence. \\
        & perception that fairness improvements in work come at personal costs. \\
        & people attributing hardships to unfair treatment by large institutions, despite limited evidence. \\
        \hline
        \multirow{6}{*}{\rotatebox{90}{\parbox{\parboxsize}{\textbf{Loyalty/\\Betrayal}}}} &
        feelings of betrayal by close connections loyal to other groups. \\
        & discussions of declining loyalty among friends based on social trends. \\
        & belief that betrayal is more common in large, organized groups than in personal circles. \\
        & loyalty within specific social or cultural groups, but only on select issues. \\
        & frustration with family members prioritizing personal ambition over traditional family values. \\
        \hline
        \multirow{6}{*}{\rotatebox{90}{\parbox{\parboxsize}{\textbf{Authority/\\Subversion}}}} &
        perception that authority is expanding, especially from non-political experts. \\
        & respect for authority figures who uphold effective traditional methods. \\
        & discomfort with religious authorities, despite a calming effect of rituals. \\
        & trust for specific authority figures (e.g., health leaders) but skepticism toward political leaders. \\
        & frustration with authority figures enforcing outdated rules. \\
        \hline
        \multirow{6}{*}{\rotatebox{90}{\parbox{\parboxsize}{\textbf{Sanctity/\\Degradation}}}} &
        discussing certain practices or traditions as sacred within specific contexts. \\
        & disgust toward perceived degradation of public spaces, despite some improvements. \\
        & outrage over misuse of religious or cultural symbols in media or fashion. \\
        & disappointment with younger generations for not valuing certain practices as sacred. \\
        & prioritizing environmental preservation locally over global concerns. \\
        \hline
        \multirow{6}{*}{\rotatebox{90}{\parbox{\parboxsize}{\textbf{Liberty/\\Oppression}}}} &
        empowerment from resisting rules perceived as unjust by large organizations. \\
        & perception of freedom being restricted by government, even for public safety. \\
        & frustration with family not recognizing personal desire for autonomy. \\
        & people feeling free in personal lives but see society becoming more oppressive. \\
        & anger toward perceived restrictions on freedom of speech or expression. \\
        \hline
    \end{tabular}
    \caption{Full list of trends, categorized by the moral foundation (MF) that was used to generate them.}
    \label{tab:trends}
\end{table*}

\begin{figure}[h!]
    \centering
    \includegraphics[width=\linewidth]{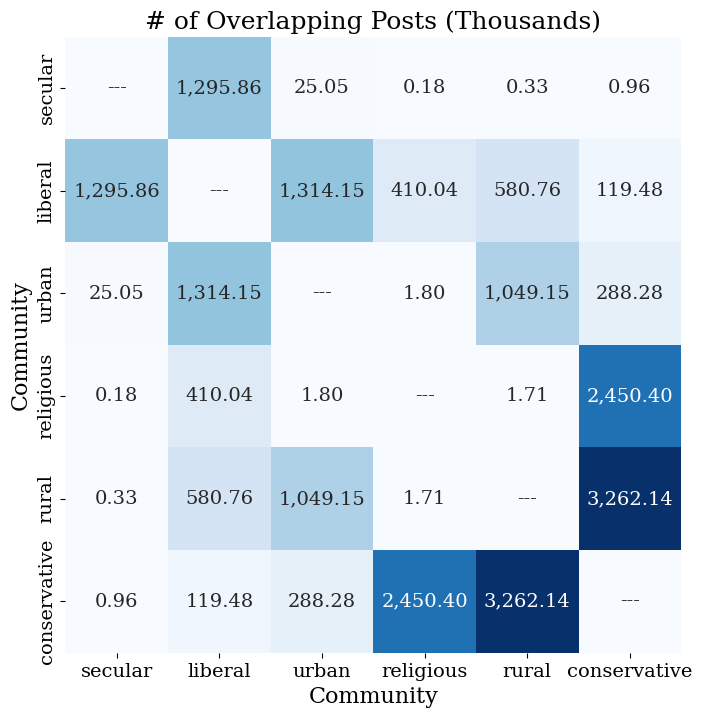}
    \caption{Number of overlapping posts in the full community datasets $D_c$. We note that there is an expected large overlap in the pairs rural/conservative, religious/conservative, liberal/secular, and liberal/urban. The less-expected overlap urban/rural is likely due to many medium-dense geographic region subreddits which the LLM labeled as both rural and urban (e.g. `r/ontario', 'r/Chattanooga', and 'r/Spokane').}
    \label{fig:overlap}
\end{figure}

\begin{figure}[h!]
    \centering
    \includegraphics[width=\linewidth]{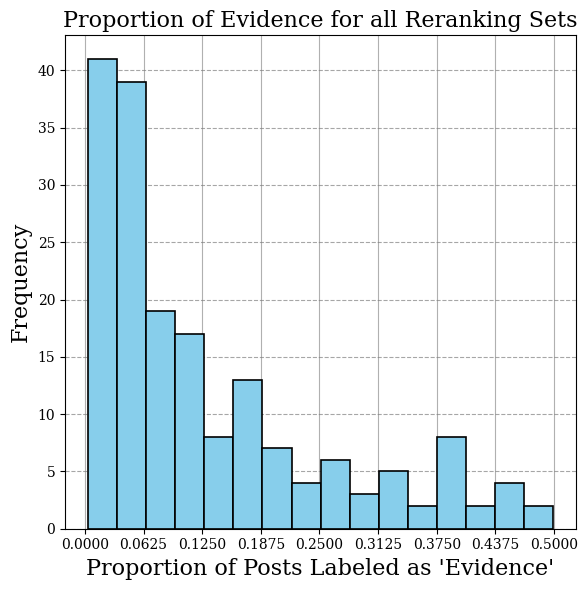}
    \caption{Proportion of posts labeled as `evidence' for each reranking set of 2000 posts. Each set corresponds to a particular trend/community pair (the subset $D_{c,q}$).}
    \label{fig:evidence-proportion}
\end{figure}

Table~\ref{tab:community_top_subreddits} contains the top 10 subreddits in each community by number of posts. Many communities have a large number of equally sized subreddits at the top, as a threshold was used to prevent one subreddit from subsuming the entire community. Thresholds were selected for each community in order to stay between 15M and 50M posts. Table~\ref{tab:trends} includes all 30 trends in the dataset. Figure~\ref{fig:overlap} shows the number of overlapping posts between each community's sub-dataset. Figure~\ref{fig:evidence-proportion} shows all proportions of posts out of 2000 with a label of `evidence'. Among 180 reranking sets, most have an evidence proportion between 2\% and 12\%.

We recruited 8 annotators with a bachelor's degree or higher to corroborate the labeling mechanism. Each annotator was shown a trend and a post, and asked to label the post's evidence level 1-5 (1 being a refutation or irrelevant, and 5 being perfect evidence). Posts were sampled randomly half and half from the top 50/2000 according to ColBERT's ranking and the bottom 1950/2000 for each community. The number of LLM-labeled `evidence' and `not evidence' posts was matched for each. Overall, 12 hours of annotation were used to double label 400 posts across 20 different trends. For each trend/post pair, two humans separately scored the evidence, and the average score was taken. Per the rubric used for scoring, posts with an average human score $\geq 3$ were binarized as `evidence' and all others were labeled `not evidence'. Among pairs of annotators, binary agreement was 70\% and the Pearson correlation of the raw scores was 0.43. Using the average binarized score as the gold label, the LLM achieved 68\% agreement with the humans and as a classifier achieved an F1 score of 0.60.

\section{Weighting}
\label{app:weighting}
The process we used for weighting was as follows: When a child concept is added to a parent, it first redistributes the weight of all siblings so that all siblings have weight $1/(\# \text{ siblings})$. Here, we consider only promoted concepts as siblings of each other and only demoted concepts as siblings of each other. Then, every child is multiplied by the product of its ancestors' weights. This step greatly reduces the children's weights with respect to their parent, grandparents, etc. Finally, the root is reset to a predefined value (we used 0.1, but did not tune this value), and the remaining weight is distributed among its children. This ensures that the root does not diminish indefinitely. Many other weighting schemes are possible and we believe this is an interesting direction for future work, but was not explored in this work.

% \clearpage
\section{Prompts}
\label{app:prompts}

\begin{figure}[h!]
    \centering
    \begin{tcolorbox}[colback=gray!30, colframe=black, boxrule=0.5mm, width=\columnwidth]
    \tiny
    I am trying to find evidence of the following trend using social media data: \{trend\}. In order to do this, I am trying to see how many posts provide evidence of this trend. Think about what kinds of things relevant people would say on social media if the trend were true. You will be given a post. Your task is to determine whether the post can be used as evidence for the trend, or if it cannot. For example, if the trend were "Increase in rural appreciation of art due to a family relative", and the post reasonably sounded like it were written by a farmer discussing a new painting hobby encouraged by his sister, then that would be evidence of the trend. Make sure to pay attention to every component of the trend when deciding if the post is evidence.\\

    Can the post be used as evidence? Clearly answer with "Yes" or "No".\\

    \#\#\# POST \#\#\#\\
    \{post\}\\

    \#\#\# ANSWER \#\#\#\\
    \end{tcolorbox}
    \caption{Prompt for labeling as evidence/not evidence.}
    \label{fig:labeling-prompt}
\end{figure}

\begin{figure}[h!]
    \centering
    \begin{tcolorbox}[colback=gray!30, colframe=black, boxrule=0.5mm, width=\columnwidth]
    \tiny
    I am trying to analyze the following trend using social media data: \{trend\}. I have a list of categories of posts. I want to know which category is best for finding evidence and which is worst.\\
    
    You will be given the list of categories. To help you know what the categories' posts are like, each category also comes with some examples of posts. Using the category name and example posts, determine the category where I am most likely to find posts that are evidence of the trend, and also determine the category where I am least likely to find such posts. Remember that my goal is to analyze the trend.\\
    
    Respond with a list of the best best categories' indices, followed by a list of the worst worst categories' indices, separated by a single line. Format your response like this:\\
    
    best\_index, second\_best\_index,...
    worst\_index, second\_worst\_index,...\\
    
    If there are no good categories or no bad categories then you can just leave a blank line for that list. Here are the categories and example posts:\\
    
    \#\#\# CATEGORY AND POSTS \#\#\#\\
    1. \{cluster1\_name\}: \{cluster1\_post1\}, \{cluster1\_post2\},...\\
    2. \{cluster2\_name\}: \{cluster2\_post1\}, \{cluster2\_post2\},...\\
    ...\\
    
    Now choose the best and worst categories and put them in the order described above. Respond only with the two lists of indices.
    \end{tcolorbox}
    \caption{Prompt for explore operation to determine supporting and refuting clusters. The top PBF and bottom DBF cluster indices are actually used.}
    \label{fig:explore-prompt}
\end{figure}

\begin{figure}[h!]
    \centering
    \begin{tcolorbox}[colback=gray!30, colframe=black, boxrule=0.5mm, width=\columnwidth]
    \tiny
    I am trying to analyze the following trend using reddit data: \{trend\}. I have a list of categories of posts. I want to know what categories are missing from my list that would provide evidence of the trend. You will be given my list of categories. To help you know what the current categories' posts are like, each category also comes with some examples of posts. Looking at the categories and example posts, come up with \{EBF\} new categories of posts and \{n\} posts per category that contain strong evidence of the trend. Remember that my goal is to get evidence of the trend.\\
    
    Given Categories and Posts:\\
    
    \{cluster1\_name\}: \{cluster1\_post1\}, \{cluster1\_post2\},...\\
    \{cluster2\_name\}: \{cluster2\_post1\}, \{cluster2\_post2\},...\\
    ...\\
    
    Now come up with the missing categories and their respective posts. Please match posts' style and length to the given posts when writing the new posts. Respond with exactly \{EBF\} new categories and \{n\} new posts for each category. Put the list of categories in this example's format, and do not include anything else in your response:\\
    
    <1st Category Description>\\
    Example Posts:\\
    "first example post for first category"\\
    "second example post for first category"\\
    "third example post for first category"\\
    ...\\
    "nth example post for first category"\\
    
    <2nd Category Description>\\
    Example Posts:\\
    "first example post for second category"\\
    "second example post for second category"\\
    "third example post for second category"\\
    ...\\
    "nth example post for second category"\\
    
    ...
    
    <mth Category Description>\\
    Example Posts:\\
    "first example post for mth category"\\
    "second example post for mth category"\\
    "third example post for mth category"\\
    ...\\
    "nth example post for mth category"\\
    \end{tcolorbox}
    \caption{Prompt for envision operation to create missing, supporting clusters.}
    \label{fig:envision-prompt}
\end{figure}

\begin{figure}[h!]
    \centering
    \begin{tcolorbox}[colback=gray!30, colframe=black, boxrule=0.5mm, width=\columnwidth]
    \tiny
    \#\#\# INSTRUCTION \#\#\#\\
    I am trying to analyze the following trend using social media posts: \{trend\}. You will be given a set of posts, and I want you to extract the core properties of the posts and concepts at play which make these posts good evidence of the trend. For example:\\
    
    \#\#\# EXAMPLE TREND \#\#\#\\
    Increase in vaping and alternative nicotine products\\
    
    \#\#\# EXAMPLE POSTS \#\#\#\\
    '''can confirm, I made a significant change in my nicotine habits a few months back, and honestly, it’s been a game-changer for me. No more of the old routine, just a clean and convenient way to manage things. I can even go about my day without anyone noticing. It’s a small change, but it’s made a huge difference in my daily routine and how I feel overall. Highly recommend giving it a try if you’re looking for an alternative.'''\\
    
    '''I had a rough time quitting smoking, but changing my nicotine intake method really helped me through it. I’m 25 and had been smoking since I was 17. I tried quitting cold turkey multiple times but always ended up going back. This new approach made it so much easier to manage cravings and slowly reduce my dependency. Plus, it’s way better for my health and social life. If you’re struggling, I’d say give this new method a shot. Sometimes, it’s just about finding the right tool for the job.\\
    
    If anyone wants to chat more about quitting smoking or exploring new approaches to nicotine, feel free to pm me. Sending good vibes and support to everyone on this journey!'''\\
    
    '''"Change is hard at first, messy in the middle, and gorgeous at the end." – Robbins\\
    
    Switching up how I consume nicotine has been exactly that for me. At first, it felt awkward and I missed the old habits, but over time, it became a new routine that’s much healthier. No more worrying about smelling like smoke or finding a place to light up. It’s definitely worth pushing through the initial discomfort for the long-term benefits.'''\\
    
    '''I decided to try something different with my nicotine consumption a while ago, and it’s been a surprising improvement. It’s a small shift, but it’s helped me cut down on smoking without too much hassle. I can handle cravings better and feel a lot healthier overall. If you’re considering making a change, this might be the solution you’re looking for. It’s been worth it for me.\\
    
    Feel free to reach out if you want to discuss more about making positive changes in your nicotine habits. We’re all in this together!'''\\
    
    '''Making the switch in how I get my nicotine was tough at first, but it’s been worth it. I was tired of the old routine and wanted something better. This new approach fits into my life so much easier, and I feel great about the change. It’s amazing how a little shift can make such a big difference. If you’re thinking about changing things up, don’t hesitate. It’s one of the best decisions I’ve made.\\
    
    Anyone looking for advice or support, feel free to pm me. Good luck to everyone on their journey!'''\\
    
    \#\#\# EXAMPLE PROPERTIES/CONCEPTS \#\#\#\\
    Switching to a new nicotine intake method\\
    Improvement in health and daily routine\\
    Reducing cravings using alternative nicotine products\\
    Explicit recommendations to others to try the new method\\
    
    \#\#\# INSTRUCTION \#\#\#\\
    Here is your trend and the set of posts.\\
    
    \#\#\# TREND \#\#\#\\
    \{trend\}\\
    
    \#\#\# POSTS \#\#\#\\
    \{posts\}\\
    
    \#\#\# INSTRUCTION \#\#\#\\
    Now, extract the core properties of the posts and general concepts at play which make these posts good evidence of the trend. Respond only with the properties/concepts and format your response exactly like the example.\\
    
    \#\#\# PROPERTIES/CONCEPTS \#\#\#\\
    \end{tcolorbox}
    \caption{Prompt for generating properties from a cluster of documents (part of Concept Induction). Not shown is another prompt for when the cluster is identified as `refuting' the trend, wherein the model is asked for properties of the posts that make them \emph{refute} the trend.}
    \label{fig:property-generation-prompt}
\end{figure}

\begin{figure}[h!]
    \centering
    \begin{tcolorbox}[colback=gray!30, colframe=black, boxrule=0.5mm, width=\columnwidth]
    \tiny
    \#\#\# INSTRUCTION \#\#\#\\
    I am trying to analyze social media posts that have certain properties. You will be given some post properties, and asked to write a set of posts that collectively fits the properties. For example, if asked for 3 posts:\\
    
    \#\#\# EXAMPLE PROPERTIES \#\#\#\\
    Switching to plant-based foods\\
    Improvement in health and energy\\
    Positive impact on the environment\\
    No mention of meat\\
    Encouraging others to try plant-based diets\\
    
    \#\#\# EXAMPLE POSTS \#\#\#\\
    I found that switching to a plant-based diet really helped with not just with regularity, but also with the size and texture of my bowel movements.\\
    
    So over a year or so I began a plant-based diet. I've been completely satisfied with every meal, never counted calories, and now I feel amazing and love the positive environmental impact.\\
    
    I'm convinced, based on research, that a plant-based diet is the way to go for my physical health, and I'm making plans to convert to that type of diet over time so that the sudden change doesn't stimulate an episode.\\
    
    \#\#\# INSTRUCTION \#\#\#\\
    Here are the set of properties. Write \{num\_groundings\}, 1-2 sentence posts that match the properties. Each post should match as many properties as possible. Respond with a line-separated list of \{num\_groundings\} short posts formatted like in the example.\\
    
    \#\#\# PROPERTIES \#\#\#\\
    \{properties\}\\
    
    \#\#\# POSTS \#\#\#\\
    \end{tcolorbox}
    \caption{Prompt for generating groundings from a set of properties (part of Concept Induction).}
    \label{fig:grounding-prompt}
\end{figure}

\section{Reproducibility}
\label{app:reproducibility}

\subsection{Dataset}
All labeling of the dataset was conducted using GPT-4o mini, while all trend creation utilized GPT-4o.

\subsection{Framework}
\begin{itemize}
    \setlength{\itemsep}{0pt}
    \setlength{\parskip}{0pt}
    \item \textbf{Retriever:} A total of 2,000 documents were consistently retrieved during the process.
    \item \textbf{Characterizer:} A root weight of 0.1 was always applied. During the envision/explore step, six centroid documents per cluster were used, and each concept was supported by exactly eight grounding quotes. A maximum of 20 clusters were ever shown at the envision/explore step. BERTopic was employed for clustering, with default parameters, leveraging HDBSCAN and sBERT. Both reranking and retrieval experiments utilized PBF and EBF at a value of 5 each, while only the retrieval experiment employed a DBF of 5. All experiments were conducted with a maximum depth of 2. All LLM calls within the Characterizer were made using GPT-4o.
\end{itemize}

\subsection{Experiments}
\textbf{Baselines:}
\begin{itemize}
    \setlength{\itemsep}{0pt}
    \setlength{\parskip}{0pt}
    \item \textbf{ColBERT:} For the ColBERT baseline, the ColBERTv2 checkpoint trained on the MS MARCO Passage Ranking task was used.
    \item \textbf{ANCE:} The publicly available RoBERTa model trained on MS MARCO was utilized.
    \item \textbf{BM25:} The Elasticsearch implementation of BM25 was used.
    \item \textbf{Query2Doc:} This method involved presenting the LLM with several few-shot examples of (trend, evidence post) pairs. The LLM was then prompted to generate an evidence post $d$ for the given trend. The generated $d$ was either concatenated with the trend or directly searched using ColBERT. We tested both search methods, and also tested with and without few-shot examples. The single-shot direct search version yielded the best performance and is the version reported in the results.
    \item \textbf{LangChain MultiQueryRetriever:} This approach used the MultiQueryRetriever to rewrite the query. Then, ColBERT was used to rerank all documents and assign scores. The scores for each document were summed to create the final ranking based on the total score.
\end{itemize}

All experiments were evaluated using BEIR \cite{thakur2021beirheterogenousbenchmarkzeroshot} on our dataset.

\section{More Qualitative Analysis}
\label{app:more-qual-analysis}
\begin{figure*}[h!]
    \centering
    \begin{minipage}[b]{0.49\linewidth}
        \centering
        \includegraphics[width=\linewidth]{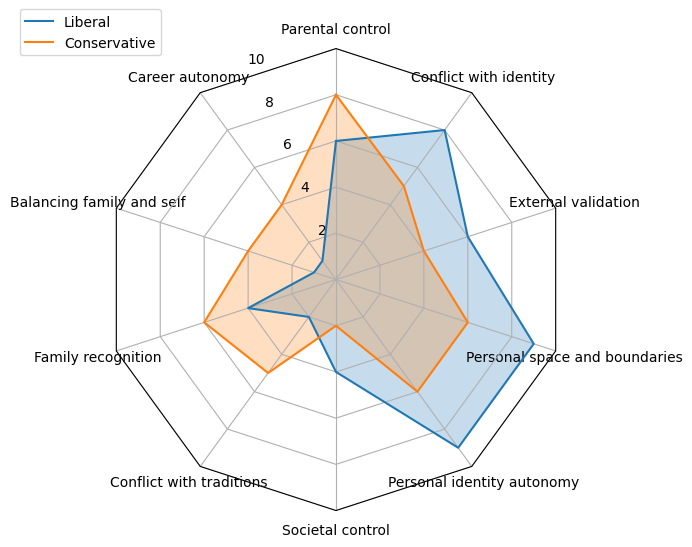}
        \caption{Trend: ``Increase in individuals expressing frustration with family members who do not recognize their personal desire for more autonomy and freedom.''}
        \label{fig:spider1}
    \end{minipage}
    \hfill
    \begin{minipage}[b]{0.49\linewidth}
        \centering
        \includegraphics[width=\linewidth]{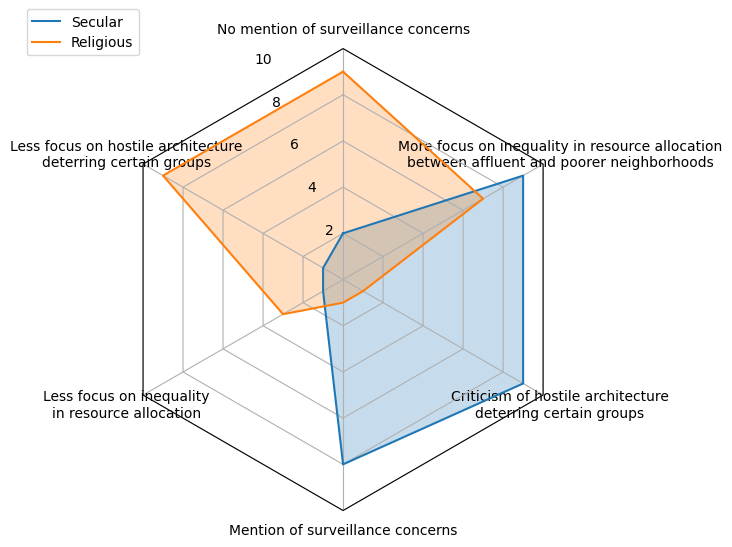}
        \caption{Trend: ``Increase in people expressing disgust toward perceived degradation of public spaces, even when some claim that conditions have improved.''}
        \label{fig:spider2}
    \end{minipage}
\end{figure*}

See Figure~\ref{fig:spider1} and Figure~\ref{fig:spider2}.

\section{Cost Analysis}
\label{app:cost-analysis}

\begin{figure*}[!htbp]
    \centering
    \includegraphics[width=\linewidth]{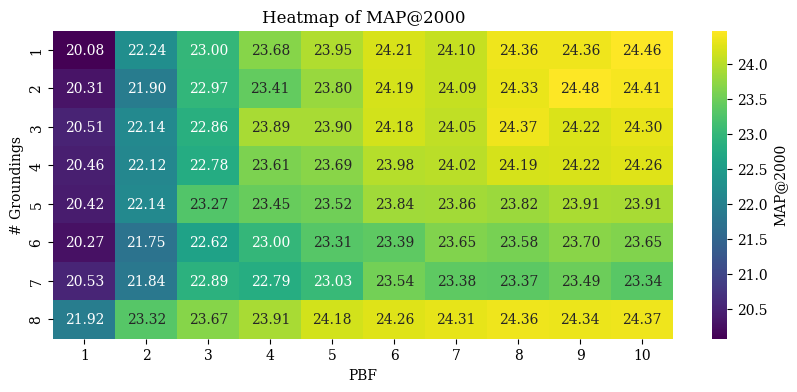}
    \caption{Effect of varying the promoted branching factor (PBF) vs. the number of groundings per concept. These were not tuned for our experiment, but show that there is a tradeoff in performance and time cost.}
    \label{fig:pbf-grounding-analysis}
\end{figure*}

In this section we discuss the cost analysis in more detail. The cost of \textsc{ConceptCarve} has two components: characterization, and retrieval. Both costs depend on several hyperparameters: the number of clusters shown during explore/envision $n$, the number of centroids shown per cluster $m$, the branching factors \(\text{EBF}\), \(\text{PBF}\), \(\text{DBF}\), and the max depth. Since marginal gains were seen after depth 2, we assume this to be the max depth in the calculations. To simplify notation, we assume that EBF, PBF, and DBF are all close, and that their sum is some general branching factor, $B$.

\textbf{Characterizer}: We calculate the cost of building the tree with total LLM input/output rather than wall time. This is because each branch of the tree can be constructed in parallel and thus sped up significantly, though this optimization is not in our implementation. Hence, we use the number of grounding-sized input/output texts, which is more akin to monetary cost when using API calls. In our case, groundings and centroid posts both have a length of about 1-3 sentences (and is therefore proportional to number of tokens).

\paragraph{Envision/Explore} In the explore step, we show $m$ clusters of $n$ documents to an LLM, which simply outputs the numbers of supporting refuting clusters. Thus the input cost is $mn$ grounding-length inputs, and output is negligible. In the envision step, the same $m$ clusters of $n$ documents are shown, but the LLM generates $B$ new sets of $n$ documents.

\paragraph{Concept Induction} In this step, clusters are first converted to properties. This is done for all supporting/refuting clusters, and the LLM is shown the $n$ centroid documents for each, so we have $B \cdot n$ grounding-length inputs, and negligible outputs (properties are much smaller than a grounding). The next step is to convert each set of properties into a set of groundings. Here the inputs are the properties (negligible), and the outputs are a set of $n$ groundings for each of $B$ clusters. Hence the result is an output of $B \cdot n$ grounding-length outputs.

These are the costs of applying the Characterizer to one concept. We do this for all non-negative concepts up to depth $2$, so we have $1$ operation on the root, and $B$ for its children. Thus the overall input cost of generating an entire tree is $(1 + B) (2m + B) (n)$, which is dominated by the terms $2Bmn + B^2n$. Likewise, the overall output cost of generating an entire tree is $(1 + B)(Bn + Bn)$, which is dominated by the term $B^2n$.

Overall, we see that the number of input tokens to the LLM scales linearly with the number of clusters shown ($m$) and the number of centroids per cluster ($n$). However, it scales quadratically with the branching factor $B$. Because we used a max depth of 2, the relationship between the total number of nodes $C$ and $B$ is $B^2 \propto C$. Hence we can say the LLM input/output tokens also scale linearly with the total number of concepts in the tree.

\textbf{Retriever}: We measure the cost of retrieval/reranking of a concept tree using the cost of retrieval/reranking of one grounding by itself. The cost of retrieval/reranking of one grounding depends totally on the standard retriever $E$ used in the backend, along with the $k$ chosen to retrieve or rerank. We denote this cost to be $E(k)$. Let $C$ be the total number of concepts in the tree, and $\gamma$ be the number of groundings per concept. Since doing retrieval/reranking on a concept tree simply does so for each grounding in each concept, the cost of a final retrieval will be $C\gamma E(k)$. In our settings, this is about $10\cdot 8$ times the cost of retrieving/reranking 2000 documents with ColBERT, thus having a latency of about 80 times that of one ColBERT retrieval. Tradeoffs in performance on the DIR task are shown in Figure~\ref{fig:pbf-grounding-analysis}, specifically between PBF and $\gamma$.

\end{document}